\documentclass[lettersize,journal]{IEEEtran}
\usepackage{amsmath,amsfonts}
\usepackage{algorithmic}
\usepackage{array}
\usepackage[caption=false,font=normalsize,labelfont=sf,textfont=sf]{subfig}
\usepackage{textcomp}
\usepackage{stfloats}
\usepackage{url}
\usepackage{verbatim}
\usepackage{graphicx}
\usepackage{hyperref}       
\usepackage{url}            
\usepackage{booktabs}       
\usepackage{nicefrac}       
\usepackage{microtype}      
\usepackage{xcolor}         
\usepackage{multirow}
\usepackage{lscape}
\usepackage{algorithm}
\usepackage{amsthm}
\usepackage{diagbox}
\usepackage{adjustbox}
\usepackage{float}
\usepackage{cite}
\usepackage{afterpage}
\usepackage{subcaption}
\usepackage{amssymb}
\usepackage{hyperref}
\usepackage{orcidlink}

\newcommand{\x}{\mathbf{x}}
\newcommand{\y}{\mathbf{y}}
\newcommand{\hy}{\hat{\mathbf{y}}}

\newcommand{\z}{\mathbf{z}}
\newcommand{\X}{\mathbf{X}}
\newcommand{\Z}{\mathbf{Z}}
\newcommand{\Q}{\mathbf{Q}}
\newcommand{\K}{\mathbf{K}}
\newcommand{\V}{\mathbf{V}}
\newcommand{\U}{\mathbf{U}}
\newcommand{\W}{\mathbf{W}}
\newcommand{\M}{\mathbf{M}}
\newcommand{\R}{\mathbb{R}}

\newtheorem{theorem}{Theorem}
\newtheorem{property}{Property}
\newtheorem{definition}{Definition}
\newtheorem{claim}{Claim}

\def\BibTeX{{\rm B\kern-.05em{\sc i\kern-.025em b}\kern-.08em
    T\kern-.1667em\lower.7ex\hbox{E}\kern-.125emX}}
\usepackage{balance}
\begin{document}
\title{WhisperRT: Turning Whisper into a Causal Streaming Model}

\author{Tomer Krichli\,\orcidlink{0009-0005-5461-9030}, Bhiksha Raj\,\orcidlink{0000-0003-0038-5513},~\IEEEmembership{IEEE Fellow}, Joseph Keshet\,\orcidlink{0000-0003-2332-5783},~\IEEEmembership{Senior Member,~IEEE}

\thanks{This work has been submitted to the IEEE for possible publication. Copyright
may be transferred without notice, after which this version may no longer be
accessible. T. Krichli and J. Keshet are with the Andrew and Erna Viterbi Faculty of Electrical and Computer Engineering, Technion--Israel Institute of Technology, Haifa 3200003, Israel (e-mail: tomerkrichli@campus.technion.ac.il; jkeshet@technion.ac.il). B. Raj is with the Language Technologies Institute, School of Computer Science, Carnegie Mellon University, Pittsburgh, PA 15213 USA.}
}

\maketitle

\begin{abstract}
Automatic Speech Recognition (ASR) has seen remarkable progress, with models like OpenAI Whisper and NVIDIA Canary achieving state-of-the-art (SOTA) performance in offline transcription. However, these models are not designed for streaming (online or real-time) transcription, due to limitations in their architecture and training methodology. We propose a method to turn the transformer encoder-decoder model into a low-latency streaming model. 

The encoder is made causal to process audio incrementally, while the decoder conditions on partial encoder states to generate tokens aligned with the available temporal context. This requires explicit synchronization between encoded input frames and token emissions. Since tokens are produced only after sufficient acoustic evidence is observed, an inherent latency arises, necessitating fine-tuning of the encoder–decoder alignment mechanism. We propose an updated inference mechanism that utilizes the fine-tuned causal encoder and decoder to yield greedy and beam-search decoding, and is shown to be locally optimal. 

Experiments on low-latency chunk sizes (less than 300 msec) show that our fine-tuned model outperforms existing non-fine-tuned streaming approaches in most cases, while using a lower complexity. We release our training and inference code, along with the fine-tuned models, to support further research and development in streaming ASR.
\end{abstract}


\section{Introduction}
\IEEEPARstart{A}{utomatic} speech recognition (ASR) systems have made remarkable progress in recent years, driven in large part by the development of powerful neural architectures. Since the introduction of the Transformer architecture \cite{vaswani2023attentionneed}, a wide range of strategies have been proposed to adapt it to ASR tasks. These include models based solely on the Transformer encoder \cite{schneider2019wav2vecunsupervisedpretrainingspeech, baevski2020wav2vec20frameworkselfsupervised, hsu2021hubertselfsupervisedspeechrepresentation}, others employ Transformer decoder-only architectures \cite{chen2024streamingdecoderonlyautomaticspeech, tsunoo2024decoderonlyarchitecturestreamingendtoend}, and some adopt the full Transformer encoder-decoder model \cite{radford2022robustspeechrecognitionlargescale, moritz2020streaming}. The most well-known model in the latter category is Whisper \cite{radford2022robustspeechrecognitionlargescale}, trained on 680,000 hours of weakly supervised speech data across 100 languages. Its broad linguistic coverage and robustness to noise have established it as a strong baseline for multilingual ASR.

In offline settings, Whisper offers near-unmatched performance in terms of noise robustness and multilingual versatility \cite{kim2024automatic}. Nevertheless, its non-causal encoder constrains its suitability for streaming applications. Prior studies \cite{wang2024simul, machavcek2023turning} have investigated deploying Whisper in a streaming context without additional training, relying instead on heuristic strategies: \emph{Simul-Whisper} \cite{wang2024simul} leverages alignment heads to determine token emission timing, whereas \emph{Ufal-Whisper} \cite{machavcek2023turning} employs an audio buffering mechanism combined with a local agreement algorithm \cite{liu2020low} to produce real-time transcriptions. Although these approaches circumvent fine-tuning, they require padding inputs to 30 seconds at every step, resulting in sub-optimal computational efficiency.

U2-Whisper \cite{zhou2025adaptingwhisperstreamingspeech} proposes fine-tuning Whisper encoder using a causal mask and trains an additional connectionist temporal classification (CTC) \cite{graves2006connectionist} head from scratch. During inference, it performs two-pass decoding, by first predicting tokens using the CTC head, then ranking candidates with Whisper decoder. WhisperFlow \cite{wang2024efficient} proposes fine-tuning the Whisper decoder to detect a fixed hush word appended to the end of each chunk, adapting Whisper to a streaming setting.
 
We introduce a method to transform the Whisper encoder-decoder architecture into a fully streaming model that operates with low latency and without dependence on future context for speech representation calculation. Unlike prior approaches that either avoid training or rely on auxiliary heads and multi-pass decoding, the proposed method directly adapts Whisper's core components for causal and computationally efficient inference. 

We propose an adaptation of Whisper to operate on a causal audio stream while preserving its established transcription quality and low word error rate (WER). The inference procedure is modified accordingly. First, the encoder is made causal, processing incoming audio incrementally as a stream rather than as a complete utterance. Under this formulation, the encoder maintains a rolling causal representation of the input. Second, the decoder conditions on this partial encoder state to generate token predictions up to the temporal boundary represented by the encoder. This requires explicit synchronization between the input audio frames encoded at a given time and the corresponding output tokens. Because token generation is performed at the granularity of complete tokens, an inherent latency arises between the encoder’s temporal coverage and the timestamp of the most recently emitted token. Once the subsequent token has been acoustically realized in the input signal, the decoder is expected to generate the corresponding prediction. Achieving this behavior necessitates careful tuning of the synchronization mechanism between encoder progression and decoder emission.

In practice, this approach is implemented by fine-tuning the model to adapt encoder representations to a causal streaming setting, together with the introduction of a time-synchronous decoding mechanism augmented by confidence-based verification. This mechanism leverages Whisper’s token-level confidence over time to address instability in the most recent predictions, which may shift as new audio chunks are incorporated and the encoder state evolves. 
Specifically, upon the arrival of each new audio chunk, a confidence measure derived from the model's output distribution is used to assess whether recently emitted tokens should be retained or revised. To support this behavior, the original non-causal encoder is modified to operate causally, and both encoder and decoder are fine-tuned using Low-Rank Adaptation (LoRA) \cite{hu2021loralowrankadaptationlarge}. Training is conducted on a weakly aligned dataset, where alignments are obtained using forced alignment tools, such as the Montreal Forced Aligner (MFA) \cite{mcauliffe2017montreal}, enabling the decoder to learn temporal boundary detection under streaming conditions. This joint adaptation of encoder and decoder allows the model to accommodate the causal representation without requiring CTC-based objectives or architectural changes.

The paper is organized as follows. Section \ref{section:realted-work} surveys prior work, Section \ref{section:problem-setup} outlines the problem setup, Section \ref{section:method} details our inference methods, Section \ref{section:training-process} details the training process,  Section \ref{section:efficiency} provides efficiency analysis, Section \ref{section:experiments} presents experimental setup, displays and discusses results, and Section \ref{section:conclusions} concludes with future directions.

For implementation details, refer to our GitHub repository\footnote{\href{https://github.com/tomer9080/WhisperRT-Streaming}{https://github.com/tomer9080/WhisperRT-Streaming}} and the Hugging Face demo page\footnote{\href{https://huggingface.co/spaces/MLSpeech/WhisperRT-low-latency-streaming}{https://huggingface.co/spaces/MLSpeech/WhisperRT-low-latency-streaming}}.

\section{Related Work}
\label{section:realted-work}

A wide range of streaming ASR methods has been proposed over the years. In this section, we restrict our focus to approaches based on Transformer architectures~\cite{vaswani2023attentionneed}, which currently define the state of the art in ASR. Moritz \emph{et al.} \cite{moritz2020streaming} combine a Transformer encoder with a CTC head and a triggered attention \cite{moritz2019triggered} decoder. Moritz \emph{et al.} \cite{moritz2021dual} propose a dual causal/non-causal self-attention mechanism in conjunction with a triggered-attention \cite{moritz2019triggered} decoder, yielding a dual ASR model. These works propose models trained from scratch and cannot easily adapt to our goal of adapting an existing model for streaming operations. Other works \cite{chen2024streamingdecoderonlyautomaticspeech, tsunoo2024decoderonlyarchitecturestreamingendtoend} investigate decoder-only Transformer architectures, which naturally fit the streaming paradigm due to the causal nature of the Transformer decoder; however, again, they do not fit our goal of adapting an existing encoder-decoder transformer to a streaming model. 

A complementary line of work focuses on augmenting model training with additional heads and auxiliary objectives. Wang \emph{et al.} \cite{wang2020low} train a Transformer encoder to detect word boundaries using forced alignment \cite{mcauliffe2017montreal}. Jia \emph{et al.} \cite{jia2024efficient} propose a SpeechLLM method that interleaves audio and text embeddings into a large language model (LLM) decoder, trained on a CTC-aligned dataset, while limiting the past context window. Choi \emph{et al.} \cite{choi2025device} fine-tune self-supervised speech models (S3Ms), exploring different masking strategies to adapt the models to streaming. Tsunoo \emph{et al.} \cite{tsunoo2021streaming} propose an encoder-decoder Transformer trained via knowledge distillation \cite{hinton2015distilling, li2014learning, lu2017knowledge}. To enhance transcription quality, they introduce block-wise synchronous beam search combined with block boundary detection techniques. Notably, these approaches modify the original model structure through additional components or training objectives.

Although the Whisper model \cite{radford2022robustspeechrecognitionlargescale} was not designed for streaming due to the non-causal nature of its encoder within the encoder-decoder framework, several studies have attempted to overcome this limitation. Macháček \emph{et al.} \cite{machavcek2023turning} avoid fine-tuning by using an audio buffer as input and emit transcriptions only after complete utterances are formed, applying local agreement \cite{liu2020low} between sub-utterances. Similarly, Wang \emph{et al.} \cite{wang2024simul} also forgo fine-tuning by developing a heuristic based on Whisper alignment heads and training a truncation detection module to determine when to emit tokens. Although these approaches do not require fine-tuning Whisper, both methods suffer from inefficiencies, as they require padding each buffer input to the full context length of the model, resulting in unnecessary computation.

Wang \emph{et al.} \cite{wang2024efficient} propose training Whisper's decoder to detect a designated hush-word at the end of each chunk, enabling the model to stop decoding appropriately. They also introduce a hybrid CPU/GPU pipeline that offloads most of the decoding process to idle CPU resources. This approach does not require input padding to the full context length for each chunk, but still performs a non-causal self-attention calculation on the encoder side. 

Zhou \emph{et al.} \cite{zhou2025adaptingwhisperstreamingspeech} propose training a new CTC head on top of the Whisper encoder, which is fine-tuned using a causal mask to make the encoder output compatible with streaming input. In this method, inference is conducted in two passes: the CTC head predicts the most probable next token, and the Whisper decoder ranks the candidate hypotheses. However, this method suggests a more complex decoding mechanism that requires both the CTC decoder and Whisper's decoder to produce a reliable transcription.

While there is a growing body of work aimed at adapting pre-trained models such as Whisper for streaming applications, existing approaches have notable limitations. \cite{wang2024simul,machavcek2023turning,wang2024efficient} do not achieve true streaming behavior, while \cite{zhou2025adaptingwhisperstreamingspeech} introduce high computational overhead. Furthermore, approaches such as those proposed by Zhou \emph{et al.} \cite{zhou2025adaptingwhisperstreamingspeech} and Wang \emph{et al.} \cite{wang2024efficient} require modifications to the model architecture or fine-tuning of the entire model, leading to a substantial increase in the number of parameters needed specifically for streaming tasks, in addition to the original Whisper model.







\section{Problem Setup}
\label{section:problem-setup}

We start by describing an encoder-decoder Transformer-based ASR model, such as Whisper \cite{radford2022robustspeechrecognitionlargescale}. It consists of a front-end composed of a few CNN layers, a transformer encoder, and a transformer decoder. The model receives the log-mel spectrogram a feature representation of speech as input. The input passes through a CNN layer that narrows the time dimension. Denote $\X_T=\begin{pmatrix} \x_1,\ldots, \x_T \end{pmatrix}$ the output sequence from the front-end, which serves as input to the transformer encoder, where $T$ is the duration of the front-end sequence, and $\x_t \in \R^d$ for $1 \le t \le T$. The encoder maps the front-end sequence to a sequence of representations $\Z_T=\begin{pmatrix} \z_1,\ldots,\z_T \end{pmatrix}$:
\begin{equation}
    \mathrm{Encoder}(\X_T)=\Z_T~,
\end{equation}
where $\z_t\in\R^d$ for $1\le t \le T$. The decoder receives as input the representations from the encoder and predicts the probability distribution over the possible set of tokens, $\mathcal{V}$, auto-regressively. We denote by $\hy_{<i}$ the vector of all previous tokens, $\hy_{<i}=(\hat{y}_1,\ldots,\hat{y}_{i-1})$. Formally, the decoder predicts the conditional probability of the next token $\hat{y}_i\in\mathcal{V}$ given all previous tokens and the representations:
\begin{equation}
    \label{eq:decoder-def}
    \mathrm{Decoder}(\textbf{Z}_T,\hy_{<i})=P(\hat{y}_i|\hy_{<i},\Z_T)~.
\end{equation}
For ease of notation, we will use the notations ${P}(y_i|\hy_{<i},\Z_T)$ and ${P}(y_i|\hy_{<i},\X_T)$ interchangeably, where they both refer to the same distribution.
In the case of non-streaming greedy decoding, the predicted token is the one that maximizes the decoder output:
\begin{equation}
\label{equation:argmax-equation}
     v^*_i = \arg\max_{v\in\mathcal{V}} P(\hat{y}_i = v|\hy_{<i},\Z_T)
\end{equation}

The non-streaming encoder operates on a fixed number of frames $T$ as input ($T=1500$). During streaming, the ASR system processes a \emph{chunk} of speech composed of $\tau$ frames, which correspond to $\tau$ vectors of the encoder representation. In other words, $\tau$ defines the resolution at which streaming occurs. More precisely, the $k$-th chunk is represented as $\X_{(k-1)\tau}^{k\tau} = (\x_{(k-1)\tau}, \x_{(k-1)\tau+1}, \ldots, \x_{k\tau - 1})$. The \emph{streaming encoder} operates on one such chunk at a time, producing the corresponding sequence of representations. Consequently, the input chunk $\X_{(k-1)\tau}^{k\tau}$ yields the output $\tilde{\Z}_{(k-1)\tau}^{k\tau}$.
As we will show, this sequence \(\tilde{\Z}_{(k-1)\tau}^{k\tau}\) differs from the subsequence \({\Z}_{(k-1)\tau}^{k\tau}\), which would result from encoding the entire front-end output \(\X_T\) and then selecting the relevant portion spanning \((k-1)\tau\) to \(k\tau\). This discrepancy motivates our use of distinct notation for these two sequences. We denote the streaming encoder for input of duration $\tau$ as follows:
\begin{equation}
    \mathrm{StreamingEncoder}({\X}_{(k-1)\tau}^{k\tau})=\tilde{\Z}_{(k-1)\tau}^{k\tau}~.
\end{equation}


The streaming decoder gets as input the hidden state subsequence $\tilde{\Z}_{(k-1)\tau}^{k\tau}$ and predicts auto-regressively the probability of the next token $\tilde{y}_{i} \in \mathcal{V}$ given the previous predicted tokens $\tilde{\y}_{<i}=(\tilde{y}_1,\ldots,\tilde{y}_{i-1})$: 
\begin{equation}
    \mathrm{StreamingDecoder}(\tilde{\Z}_{(k-1)\tau}^{k\tau},\tilde{\y}_{<i})={P}(\tilde{y}_{i}|\tilde{\y}_{<i},\tilde{\Z}_{(k-1)\tau}^{k\tau})~.
\end{equation}


Our goal is to adapt the existing Whisper encoder and decoder to streaming tasks. Both the encoder and the decoder will be fine-tuned with the objective that the overall performance of the streaming ASR will be comparable to that of the offline ASR. Formally, let $\X_{k\tau}$ be the input up to frame $k\tau$ and denote by $\y_{k\tau}$ the corresponding ground truth token sequence. Similarly, denote $\hat \y_{k\tau}, \tilde \y_{k\tau}$ the predicted token sequence of the offline and the streaming decoder respectively. We aim that the word error rate (WER) of the streaming decoder $\mathrm{WER}(\tilde \y_{k\tau},\y_{k\tau})$ is low and comparable to $\mathrm{WER}(\hat \y_{k\tau}, \y_{k\tau})$.

\section{Method}
\label{section:method}

Originally, the transformer encoder utilizes non-causal self-attention mechanism \cite{vaswani2023attentionneed}. The encoder is composed of $L$ self-attention (SA) layers. Each SA layer $l$ receives an input $\U^{l-1}_T$ from the previous layer $l-1$, where $\U^0_T= \X_T$, and $T$ is the total number of frames in the input. It outputs the queries $\Q\in \R^{T\times d}$, the keys $\K\in \R^{T\times d}$, and values $\V\in \R^{T\times d}$, where $d$ denotes the vector dimension. Specifically, the queries, keys, and values are all calculated using the same $\U_T$:
\begin{equation}
\Q^l_T=\U^{l-1}_T\W^l_Q, ~~~ \K^l_T=\U^{l-1}_T \W^l_K,~~~ \V^l_T=\U^{l-1}_T \W^l_V~,
\end{equation}
where $\W^l_Q, \W^l_K, \W^l_V \in \R^{d \times d}$ are learned matrices for layer $l$. The \emph{self-attention} is defined as
\begin{equation}\label{eq:SA}
\mathrm{SA}^l(\U^{l-1}_T)=\mathrm{Softmax}\left(\frac{\Q^l_T{\K^l_T}^{\top}}{\sqrt{d}}\right) \V^l_T ~.
\end{equation}
After the self-attention layer, a feed-forward (FF) layer is then applied using a residual connection. 


\begin{figure}
    \centering
    \includegraphics[width=0.65\linewidth]{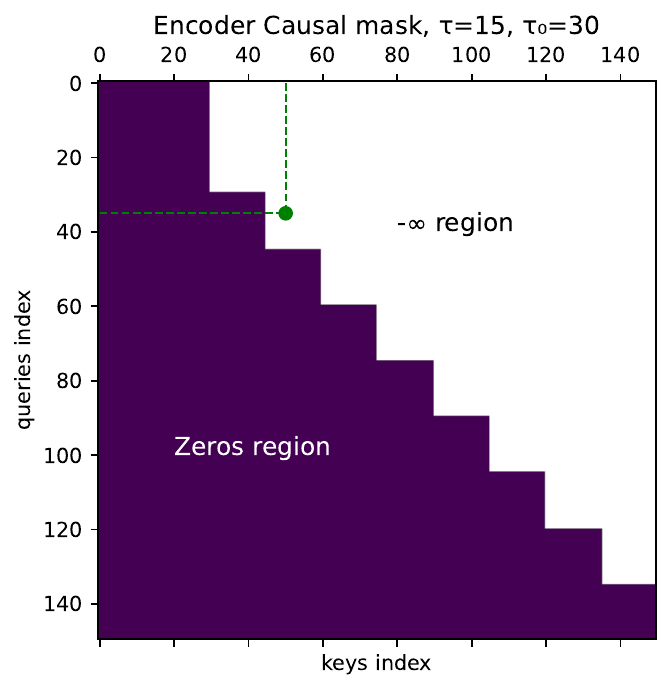}
    \caption{Encoder causal mask example, $\tau=15, \tau_0=30$ given $k=10$ chunks. Such mask applies that the model waits $600$ msec for the first buffer before feeding the input to the encoder. Then, input is being fed every $300$ msec. Purple regions contain zeros while white regions contain $-\infty$. The index (35,50) is marked in a green point.}
    \label{fig:mask}
\end{figure}

\subsection{Streaming Encoder}

Working in a streaming setting means that a \emph{mask matrix} is applied to the full self-attention matrix to ``prevent'' access to future input frames. Elements in the matrix that correspond to no masking are assigned to be 0 whereas the elements that are masked are assigned the value of $-\infty$. We define two parameters of the masking. Let $T$ be the maximal context available on the encoder's input. Let $\tau < T$ be the chunk size or the granularity of encoder frames, and $k\in\mathbb{N}$ the chunk index. In our setting the first chunk is larger than subsequent chunks, and $\tau_0 < T$ denotes its size.

We exemplify this notation using Figure ~\ref{fig:mask}, where we assume a chunk size of $\tau = 15$ frames, an initial chunk size of $\tau_0 = 30$ frames, and that the model is currently processing chunk number $k = 10$. The mask $\M(k, \tau,\tau_0)\in\{0,-\infty\}^{k\tau\times k\tau}$ is applied to the dot product in the following way:
\begin{equation}\label{eq:sa_mask}
{\mathrm{\widetilde{SA}}}(\X_{k\tau}, \tau,\tau_0)=\mathrm{Softmax}\left(\frac{\Q_{k\tau}\K_{k\tau}^{\top} + \M(k,\tau,\tau_0)}{\sqrt{d}}\right) \V_{k\tau}
\end{equation}
Here we omitted the layer super-script, and \eqref{eq:sa_mask} refers to any of the encoder layers. With this masked self-attention mechanism, the encoder's representation of the first $k\tau$ frames remains unchanged whether processing is performed chunk by chunk up to chunk $k$ or over the entire input sequence at once, as formalized in Property 2 in the supplementary material.

\subsection{Streaming Decoder}

The streaming decoder receives the encoded speech representation of the last chunk $\tilde{\Z}_{(k-1)\tau}^{k\tau}$ and the previous predicted tokens $\tilde \y_{<i}$. Its goal is to predict the distribution of the next token as in \eqref{eq:decoder-def}. To achieve this, the decoder employs a cross-attention layer. While cross-attention operates using the same mechanism as self-attention, it differs in the way queries, keys, and values are constructed. Specifically, the queries are derived from the output of the decoder self-attention layers up-to the previous token, denoted as $\bar\U^l_{<i}$, while the keys and values come from the acoustic feature representations $\tilde{\Z}_{k\tau}$. Thus, a cross-attention layer is defined as follows:
\begin{equation}
\Q^l_{<i} = \bar\U^{l-1}_{<i} \bar{\mathbf{W}}^l_Q, \quad \K^l_{k\tau} = \tilde{\Z}_{k\tau} \bar{\mathbf{W}}^l_K, \quad \V^l_{k\tau} = \tilde{\Z}_{k\tau} \bar{\mathbf{W}}^l_V 
\end{equation}
where $\bar{\W}^l_Q, \bar{\W}^l_K, \bar{\W}^l_V \in \R^{d \times d}$ are learned matrices for layer $l$ of the decoder. The \emph{cross-attention} is defined as
\begin{equation}\label{eq:CA}
\mathrm{CA}^l(\bar\U^{l-1}_{<i}, \tilde{\Z}_{k\tau}) = \mathrm{Softmax}\left(\frac{\Q^l_{<i} {\K^l_{k\tau}}^\top}{\sqrt{d}}\right) \V^l_{k\tau} 
\end{equation}
Overall, the decoder is built from a self-attention layer, a cross-attention layer and a feed-forward layer with residual connections.


\begin{figure}[t]
    \centering
    \includegraphics[width=1\linewidth]{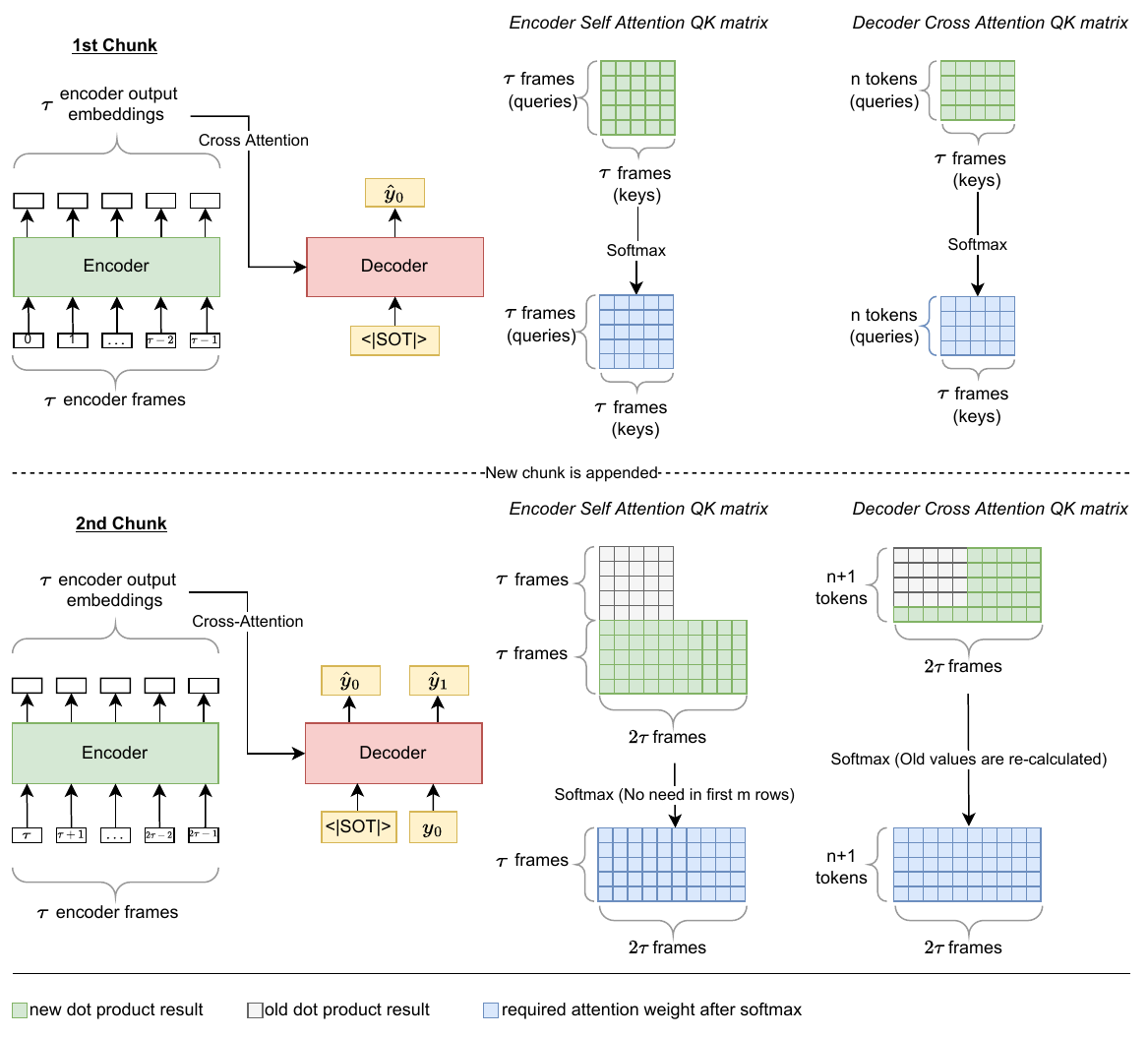}
    \caption{The inference process, using a chunk size of $\tau$, initial chunk of size $\tau_0=\tau$. The figure also illustrates how attention weight matrices are computed, specifically, within the encoder's self-attention and the decoder's cross-attention mechanisms.}
    \label{fig:attention-calculation}
\end{figure}

\subsection{Streaming Inference}
\begin{figure*}
    \centering
    \includegraphics[width=1\linewidth]{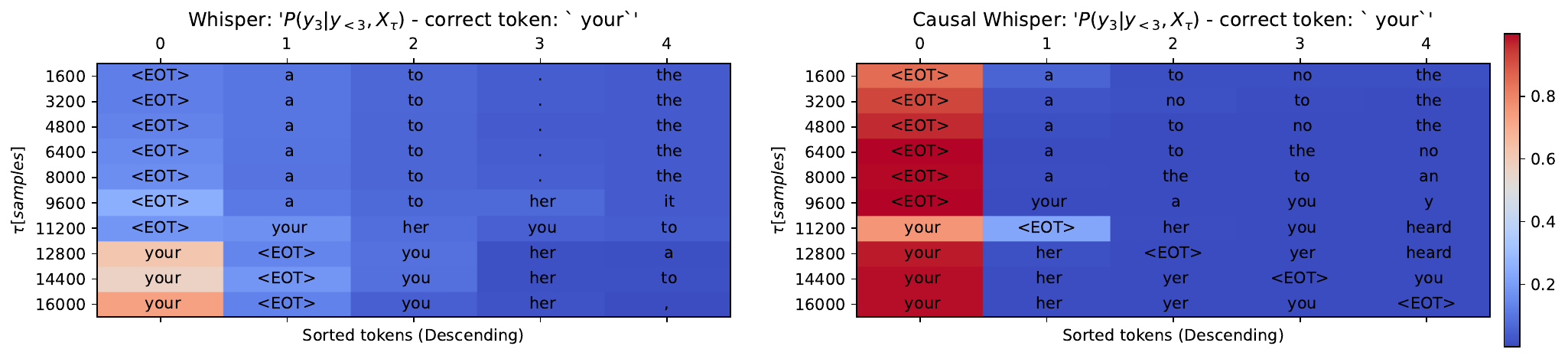}
    \caption{Token  distribution for the Whisper model (left) and WhisperRT (right) for third token over time, conditioned on the ground truth prefix `\textit{she had}.' The right color bar indicates the confidence scale for both models, with red regions representing higher confidence values. The full utterance is: ``\textit{she had your dark suit in greasy wash water all year}.'' The plots show the probability distribution of the \textbf{third} token, rather than the sequence of predicted tokens. As such, \emph{EOT} is the predicted token until there is sufficient acoustic evidence to predict `\textit{your}.' Hence, the token `\textit{had}' does not appear before the token `\textit{your}.'}
    \label{fig:models-conf-comparison}
\end{figure*}
The heart of our algorithmic novelty lies in the inference mechanism. Decoding the optimal token sequence is typically framed as predicting the most probable token
\begin{equation}\label{eq:next_token}
     v^*_i = \arg\max_{v\in\mathcal{V}} P(\hat{y}_i = v|\hy_{<i},\Z_T)~.
\end{equation}
Throughout this paper, we primarily use the \emph{max} operation to represent \emph{greedy decoding}, where the token with the highest conditional probability is selected at each step. While it is possible to \emph{sample} tokens from the probability distribution instead, this alternative is only considered in the context of \emph{beam search} decoding.

Applying the inference procedure \eqref{eq:next_token} to the streaming setting presents two primary challenges, especially when processing fixed-size input chunks that may not align with token boundaries. First, a token may be \emph{partially} contained within an input chunk, with the remainder spanning into the next one. In such cases, the model lacks sufficient information to confidently predict the complete token. Second, the chunk may not provide enough contextual information, leading to premature or unstable token predictions that would differ if future input were available, especially for tokens with similar phonetic content. To overcome these obstacles, we propose a decoding mechanism that supports token-level regression, allowing the decoder to revise past predictions in light of new information. 

To motivate our approach, consider Figure~\ref{fig:models-conf-comparison}, which illustrates the most probable tokens with temporal alignment. The X-axis shows the top-5 token predictions, while the Y-axis denotes the number of processed acoustic frames. The fine-tuned \emph{WhisperRT} model (right) exhibits increasing confidence in the correct token as more acoustic context becomes available, in contrast to the vanilla Whisper model (left). Notably, the fine-tuned model predicts the correct token \texttt{your} one frame earlier than the vanilla model.

A token is considered \emph{stable} if its prediction is consistent across two consecutive input chunks. When this condition is met, the prediction is considered \emph{finalized}. In contrast, \emph{unstable} tokens, those with inconsistent predictions are discarded, and decoding is resumed from the first unstable position. We describe this heuristic in the context of both greedy decoding and beam search decoding.

\medskip

\subsubsection{Greedy Streaming Decoding}
Greedy decoding is a decoding strategy in which, at each time step, the model selects the token with the highest predicted probability, without considering alternative candidates. We start with a definition of a \emph{stable token} for greedy decoding.
\begin{definition}[Stable token for greedy decoding]
A token $y_i = v$ predicted in the $k$-th chunk \emph{stable under greedy decoding} if it satisfies at least one of the following conditions. Either the token $v$ has a greater or equal probability given a new chunk relative to the previous chunk:
\begin{equation}
    P(y_i = v \mid \y_{<i}, \X_{k\tau}) \ge P(y_i = v \mid \y_{<i}, \X_{(k-1)\tau}) \label{eq:stability_likelihood}
\end{equation}
or $v$ is the most probable token given the new chunk:
\begin{equation}
    v = \arg\max_{u \in \mathcal{V}} P(y_i = u \mid \y_{<i}, \X_{k\tau}) \label{eq:stability_argmax}~.
\end{equation}
\end{definition}
That is, a token is considered stable as long as it remains the most probable token in the distribution. Additionally, a token may be marked as stable even if it is not the most probable token, but its probability increases along time. 

The greedy decoding algorithm is modified as follows: upon the arrival of a new chunk $k$, the prediction of token $i$ proceeds only if all of the last $n$ tokens have been verified as stable. Otherwise, the decoding process is resumed from the index of the first unstable token, discarding any subsequent tokens. The complete greedy streaming decoding process is presented in Algorithm~\ref{alg:greedy-decoding}.

The next claim states that our proposed greedy decoding will have at least a locally optimal token prediction, given that the probability $P(y_i\mid \y_{<i}, \X_{k\tau})$ was correctly estimated.
\begin{claim}[Streaming greedy decoding optimality]
Let 
\begin{equation}
P^*(y_i\mid \y_{<i}, \X_{k\tau})
\end{equation}
be the optimal probability distribution of the $i$-th token given the preceding tokens, and the acoustic data till $k\tau$. Denote by $\rho_k$ the \emph{probability path} of the predicted token sequence until the $k$-th chunk. Our proposed greedy decoding algorithm, as depicted on Algorithm~\ref{alg:greedy-decoding}, is optimal in the sense that it yields a probability path $\rho_k^{\text{CW}}$ which is larger or equal on the probability path of the greedy algorithm, $\rho_k^{\text{G}}$.
\label{theorem:greedy-optimality}
\end{claim}
The claim is proven by induction and is included in the supplementary material.

\medskip

\subsubsection{Beam-Search Streaming Decoding}

Beam search decoding is a heuristic search algorithm that explores multiple hypotheses at each time step, maintaining the top-scoring sequences within a fixed-size beam to balance between accuracy and computational efficiency. We enhance the beam search decoding process by adding a stability check on the last $n$ tokens when a new chunk arrives. Let $b$ denote the beam size. A token is considered stable if and only if it remains within the top-$b$ candidates (i.e., within the beam) after incorporating the latest acoustic frame. We define the operator $\mathrm{TopK}$ as follows.
\begin{definition}[Top-$k$ operator]
Let $\mathbf{p} \in \mathbb{R}^{|\mathcal{V}|}$ be a vector of real-valued scores indexed over a finite set $\mathcal{V}$. The \emph{Top-$k$ operator}, denoted by $\mathrm{TopK}(\mathbf{p}, k)$, returns the set of indices corresponding to the $k$ largest values in $\mathbf{p}$:
\begin{equation}
\mathrm{TopK}(\mathbf{p}, k) = \left\{ j \in \mathcal{V} \,\middle|\, j \text{ is among the } k \text{ largest entries of } \mathbf{p} \right\}.
\end{equation}
\end{definition}
Now we can define a stable token for the beam search decoding.
\begin{definition}[Stable token for beam search decoding]
A token $y_i=v$ is defined \emph{stable under beam search decoding} if it satisfies:
\begin{equation}
\label{eq:beam-stable-condition}
 v \in \mathrm{TopK}\left(P(y_i \mid \y_{<i}, \X_{k\tau}), b\right)\\
\end{equation}
and
\begin{equation}
    P(y_i = v \mid \y_{<i}, \X_{k\tau}) \ge P(y_i = v \mid \y_{<i}, \X_{(k-1)\tau}) \label{eq:stability_likelihood_beam}
\end{equation}
That is, the token is stable if it remains within the top-$b$ most probable candidates at its position in the current beam step, and its probability gradient is non-negative.
\end{definition}

Unlike greedy decoding, where a single prediction path is followed, beam search maintains multiple hypotheses in parallel. Typically, decoding is considered complete only once a sufficient number of hypotheses have terminated, i.e., their final token is the end-of-transcription (\emph{EOT}) symbol. However, in the streaming setting, this approach can introduce errors such as hallucinations or repeated phrases. This often occurs because the input buffer may end mid-word, leading some beams to favor repetitive completions or hallucinated content rather than emitting \emph{EOT}. To mitigate this, we adopt a greedy-style stopping criterion: if any hypothesis within the beam predicts an \emph{EOT} token, the decoding is paused until the next acoustic frame is received.

\begin{algorithm}[t]
\caption{Streaming Greedy Decoding}
\label{alg:greedy-decoding}
\begin{algorithmic}[1]
\STATE \textbf{Input:} $n$
\STATE \textbf{Output:} $\hat\y$

\STATE $\hat\y\leftarrow\varnothing$
\STATE completed $\leftarrow$ False
\FOR{$k\in\{0,\ldots,\frac{T}{\tau}\}$}
\STATE new\_chunk $\leftarrow$ True
\WHILE{not completed}
\IF{new\_chunk}
\STATE new\_chunk $\leftarrow$ False

// \textit{Check for unstable tokens}

  \FOR{$m\in\{n-1, n-2, \ldots, 0\}$}
      \IF{token $y_{i-m}$ is unstable}
          \STATE $\hat\y \leftarrow \hat\y_{<i-m}$
          \STATE \textbf{break}
        \ENDIF
    \ENDFOR
\ENDIF

// \textit{Decode greedily}

\STATE $\hat\y \leftarrow \hat\y_{<i-m} \oplus{\mathrm{arg}\max}~{P}(\hat y_{i-m}\mid\hat\y_{<i-m},\X_{k\tau})$
\STATE completed $\leftarrow EOT \in \hat\y$
\ENDWHILE
\ENDFOR
\STATE \textbf{return} $\hat\y$
\end{algorithmic}
\end{algorithm}

\begin{algorithm}[t]
\caption{Streaming Beam Search Decoding}
\label{alg:beam-search-decoding}
\begin{algorithmic}[1]
\STATE \textbf{Input:} $n, b$
\STATE \textbf{Output:} $\hat\y$

\textit{ // Initialize beam}
\STATE $\mathcal{B}\leftarrow \varnothing $
\STATE completed $\leftarrow$ False
\FOR{$k\in\{0,\ldots,\frac{T}{\tau}\}$}
\STATE new\_chunk $\leftarrow$ True
\WHILE{not completed}
\FOR{$\y \in \mathcal{B}$}
    \IF{new\_chunk}
        \STATE new\_chunk $\leftarrow$ False
        
        \textit{ // Check for unstable tokens}
        
        \FOR{$m \in \{n-1, n-2, \ldots, 0\}$}
            \IF{token $y_{i-m}$ is unstable}
                \STATE $\hat\y \leftarrow \y_{<i-m}$
                \STATE \textbf{break}
            \ENDIF
        \ENDFOR
    \ENDIF
    
     \textit{ // Decode with beam-search}

    \FOR{$t \in \mathrm{TopK}\left(\log P(\hat y_{i-m}~|~\hat\y_{<i-m}, \X_{k\tau}), b\right)$ } 
        \STATE $\tilde\y \leftarrow \hat\y\oplus t$
        \STATE $\mathcal B \leftarrow \mathcal B \cup \tilde \y$
    \ENDFOR
\ENDFOR

// \textit{Store in beam top $b$ hypotheses}

\STATE $\mathcal{B}\leftarrow \mathrm{TopK}(\mathcal{B}, b)$
\STATE completed $\leftarrow$ EOT $\in \mathcal{B}$
\ENDWHILE
\ENDFOR
\STATE \textbf{return} $\y$ - the most probable token sequence in $\mathcal{B}$
\end{algorithmic}
\end{algorithm}

The full beam search algorithm is available on Algorithm \ref{alg:beam-search-decoding}. Note that the suggested version of beam search decoding involves a varying length of beams during the decoding process, since the stability check per token might yield different regression indices. We maintain this beam set $\mathcal{B}$ by adding padding tokens, that their probability is not being examined when we check for the stability condition on the arrival of a new chunk to the model. Using such beam set allows us to stop the search when \textit{EOT} is in the beam, while keeping relevant beams of different lengths.  




\section{Fine-tuning Process}
\label{section:training-process}

\begin{figure*}
    \centering
    \includegraphics[width=0.9\textwidth]{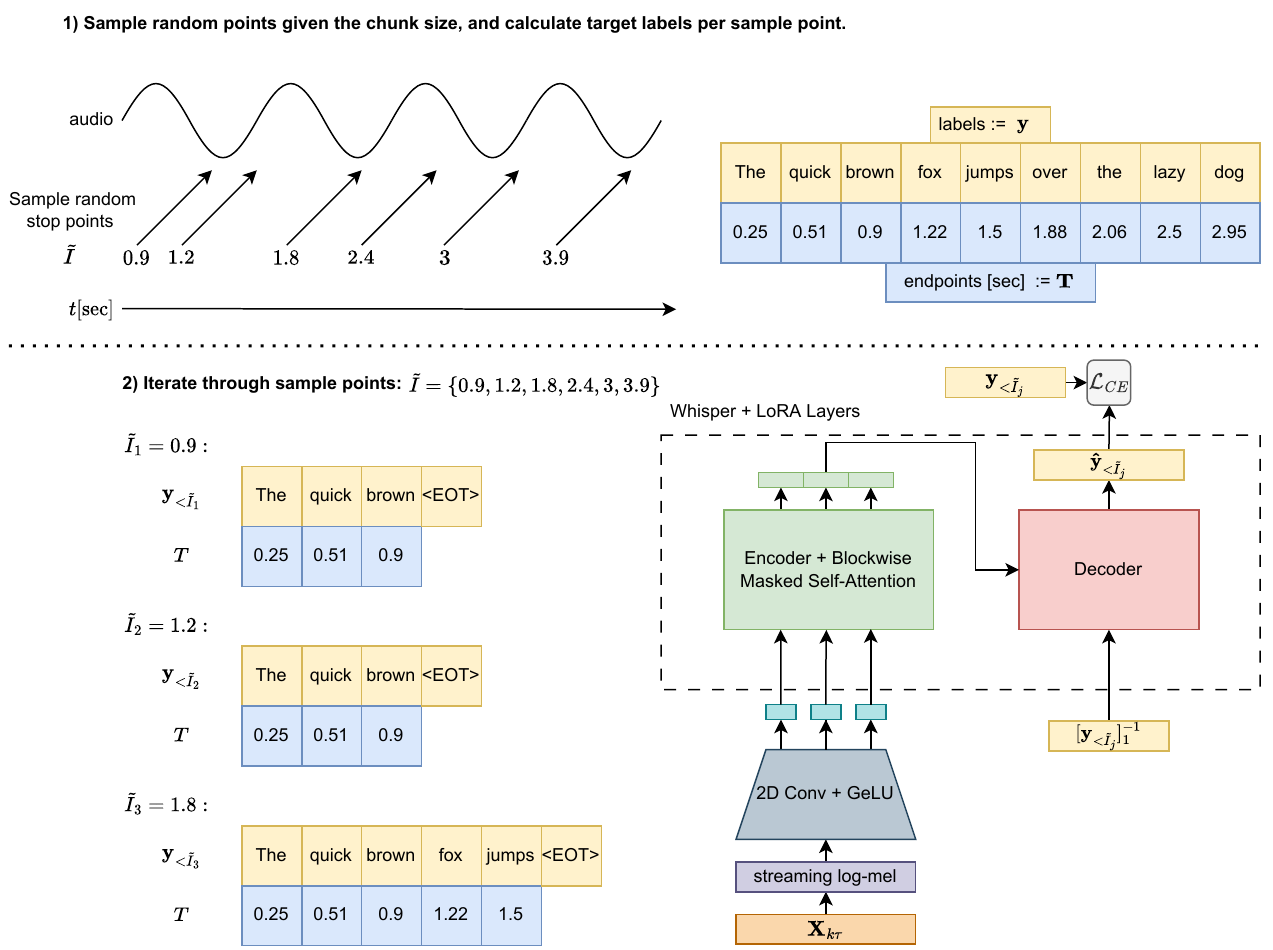}
    \caption{Illustration of the fine-tuning process. The example depicts an encoder operating with a chunk size of $300$ msec. This approach improves training efficiency by avoiding the need to process every possible frame in the streaming setting.}
    \label{fig:training-process}
\end{figure*}

To support the streaming encoder, streaming decoder, and the proposed streaming inference procedure, the model must be fine-tuned. The process is described below and illustrated in Fig.~\ref{fig:training-process}. We define two primary training objectives: (i) adapting the encoder to produce a causal speech representation, and (ii) training the decoder to determine when to emit a new token. To achieve these objectives, we integrate LoRA layers \cite{hu2021loralowrankadaptationlarge} into the encoder’s self-attention layers, as well as into the decoder’s self-attention and cross-attention layers. This design ensures that only the LoRA parameters are updated during training, yielding a compact and efficient module that supports both offline and streaming transcription. The complete fine-tuning procedure is detailed in Algorithm~\ref{alg:training-whisper-lora}.

To facilitate efficient fine-tuning, we fix a chunk size $\tau \in \mathbb{N}$ such that $0 < \tau \leq T$, where $T$ denotes the maximum encoder context length. The self-attention mask is defined based on this chunk size $\tau$, and an initial chunk size $0 < \tau_0 \leq T$ is used. Note that while sharing the same mask across all fine-tuning steps increases efficiency, it limits the model's generalization to other chunk sizes. 

A naive adaptation strategy involves evaluating all possible time points at intervals of $20\tau$ msec, denoted by the set $I$, computing the corresponding target labels using the weak alignment generated via force alignment \cite{mcauliffe2017montreal}, and back-propagating the loss. To address the inefficiency of enumerating all possible alignments, we sample only a subset of the time points in $I$. Specifically, for each fine-tuning step, we define a sampling fraction $\hat f \in (0, 1]$ and randomly select $\hat f \cdot |I|$ time points from $I$, denoted as $\tilde{I}$. We then iterate over this sampled set $\tilde{I}$ during the fine-tuning step. At each fine-tuning sub-step $j$, we compute the corresponding target sequence $\y_{<\tilde I_j}$ as:
\begin{equation}
\y_{<\tilde I_j} = \{y_i \mid t_{\text{end}}(y_i) \leq \tilde{I}_j ~ \text{sec} \}
\end{equation}
Here, $t_{\text{end}}(y_i)$ is the end time of token $y_i$, determined from the forced alignment. For each sub-step, the model is optimized using a cross-entropy loss, $\mathcal{L}_{CE}$, computed over $\y_{<\tilde I_j}$ and $\hat\y_{<\tilde I_j}$. 

Finally, we introduce a random chunk-size masking strategy during fine-tuning. While the proposed procedure effectively trains a streaming model for a fixed chunk size $\tau$, it limits inference to that specific duration. To overcome this limitation, we adopt a random chunk-size masking scheme, in which chunk sizes are uniformly sampled between 0.1 and 1.0 seconds during training. This improves the model’s ability to generalize across varying chunk sizes at inference time. We denote $\bar f$ as the number of sub-sequences within a batch used for training at each step.

\begin{algorithm}
\caption{Streaming Whisper LoRA Fine-Tuning Process}
\label{alg:training-whisper-lora}
\begin{algorithmic}[1]
\STATE \textbf{Input:} Whisper NN - $\mathcal{M}^{LoRA}$, Dataset - $\mathcal{D}$, chunk size $\tau$, initial chunk size $\tau_0$, sample points fraction $\hat f$.
\FOR{$\mathcal{B}$ in $\mathcal{D}$}
    \STATE $\X, \y \leftarrow \mathcal{B}$ 
    \STATE $\tilde{I} \leftarrow \mathrm{getSamplePoints}(\tau, \tau_0,\hat f)$
    \FOR{$j$ in $\{0,1,\ldots,|\tilde{I}|-1\}$}
        \STATE $\hy \leftarrow \mathcal{M}^{LoRA}(\X_{\tilde{I}_j},\y_{<j}, \tau, \tau_0)$
        \STATE $\ell\leftarrow \mathcal{L}_{CE}(\hy, \y)$
        \STATE calculateGradients($\ell$)
        \STATE optimizerStep()
    \ENDFOR
\ENDFOR
\end{algorithmic}
\end{algorithm}

\section{Efficiency}\label{section:efficiency}

Before turning to empirical evaluation of our proposed method we present efficiency consideration and implementation of KV-cache both for the encoder and the decoder. We then state the complexity of the proposed algorithm.

\subsection{KV-Cache: Encoder} \label{section:encoder-kv-cache}

When using causal self-attention, the encoder's computation can be optimized through a KV-cache mechanism. Because new acoustic frames depend only on the current chunk and past inputs, the key and value matrices ($\K$, $\V$) at each layer can be cached rather than recomputed at every iteration—similar to the caching strategy used in causal decoders.

By design, the output of the encoder at position $t$ is influenced only by the $t$-th row of the self-attention weights matrix. When applying a causal mask, each encoder output at index $t$ is conditioned solely on the current and preceding chunks. As a result, previously computed matrices $\K$ and $\V$ remain valid for future steps and can be reused. This caching mechanism reduces the encoder's computational complexity from quadratic to linear, significantly improving efficiency during streaming inference. A detailed complexity analysis is provided in Section~\ref{section:inference-complexity}.

\subsection{KV-Cache: Decoder}

The decoder incorporates two distinct attention mechanisms: self-attention and cross-attention, each with its own KV-cache policy. 

\paragraph{Cross-Attention KV-Cache}
As discussed in Section~\ref{section:encoder-kv-cache} and formalized in Property 2 in the supplementary material, encoder outputs corresponding to past acoustic frames are independent of future inputs. Therefore, when a new audio chunk arrives, the decoder computes $\K$ and $\V$ values only for this new chunk. Previously processed chunks are stored and reused. The dot product between the new query and the cached key-value pairs is then calculated as usual.

\paragraph{Self-Attention KV-Cache}
In contrast to cross-attention cache, caching self-attention activations in the decoder is less straightforward. This is due to the dynamic nature of the decoder's input embeddings: each buffer update introduces new acoustic information, which alters the decoder’s context and thus the token embeddings.

This distributional drift arises because the decoder continuously receives new audio data. Each decoder layer depends on a causal self-attention block, a cross-attention block, and a feed-forward network. While the self-attention is strictly causal (only attending to past tokens), the cross-attention is not: decoded tokens may attend to future acoustic frames.


Consequently, applying a self-attention KV-cache in the decoder, as done in offline settings, is no longer valid. To address this, we introduce an approximation: from some time index $k\tau < T$, we assume that $\K_{k\tau}^T, \V_{k\tau}^T$ contribute negligibly to subsequent cross-attention computations. Beyond this point, token embeddings are considered stable and can be cached. However, this assumption often fails in practice, as many cross-attention heads are not temporally aligned or monotonic.

Despite this, we propose to empirically examine this approximation to assess whether our fine-tuning method induces more monotonic or stable behavior in the model’s attention patterns.

\subsection{Complexity Analysis}
\label{section:inference-complexity}
Since the proposed streaming Transformer model does not rely on non-causal attention, we conduct a complexity analysis to estimate its computational efficiency advantages.

\begin{theorem}
Let $T$ be the input sequence length to the encoder, $d$ the embedding dimension, and $\tau$ the chunk size, with $0 < \tau \ll T$. The computation of blocked causal attention over the full sequence during streaming requires $\mathcal{O}(T^2d + Td^2)$ operations and $\mathcal{O}(Td)$ additional memory.
\label{theorem:encoder-complexity-analysis}
\end{theorem}
The proof is provided in the supplementary material. In contrast, methods such as \cite{machavcek2023turning,wang2024simul}, which do not fine-tune Whisper, pad the input signal to the maximum context size whenever a new frame arrives. As a result, the encoder must perform $\mathcal{O}(T^2d + Td^2)$ operations for each chunk \cite{vaswani2023attentionneed}. Given $\frac{T}{\tau}$ chunks, the total encoder-side computational cost becomes $\mathcal{O}\left(\frac{T^3d}{\tau} + \frac{T^2d^2}{\tau}\right)$ operations. This difference is expected to have a significant impact, especially in low-latency scenarios where the chunk size is small ($\tau \ll T$), and when using a larger model where $d$ is larger. A further demonstration of the runtime advantage appears in Section~\ref{section:experiments-runtime}.

\section{Experiments}
\label{section:experiments}

We evaluate \emph{WhisperRT} across several tasks. First, we focus on its primary objective: streaming transcription, assessed using word error rate (WER) and related variants, which we detail below. We then analyze runtime performance using metrics such as Real-Time Factor (RTF) and average latency. 

\subsection*{Evaluation Metrics}

To better assess streaming transcription quality, we employ metrics that are more suitable for streaming scenarios than the standard Word Error Rate (WER).

\paragraph{Relative Word Error Rate (RWER)}
Let $\hat{\mathbf{y}}_{\tau}$ be the hypothesis generated by the streaming model up until time point $\tau$, containing $\tilde{N}_\tau$ words. Define the set $\hat{\mathcal{Y}} = \{\hat{\mathbf{y}}_0, \ldots, \hat{\mathbf{y}}_T\}$ as the set of hypotheses generated throughout the decoding process. Let $\mathbf{y}$ be the ground-truth transcription, consisting of $N$ words, and let $\y_{\tilde N_\tau}$ be the prefix of the ground-truth string up to the $\tilde{N}_\tau$-th word. Finally, let $I$, $D$, $S$, and $C$ denote insertions, deletions, substitutions, and correct predictions, respectively. Then, RWER is defined as:
\begin{equation}\nonumber
\mathrm{RWER}(\mathbf{y}, \hat{\mathcal{Y}}) \!=\! \frac{\sum_\tau I(\y_{\tilde N_\tau}, \hat{\mathbf{y}}_\tau) + D(\y_{\tilde N_\tau}, \hat{\mathbf{y}}_\tau) + S(\y_{\tilde N_\tau}, \hat{\mathbf{y}}_\tau)}{\sum_\tau C(\y_{\tilde N_\tau}, \hat{\mathbf{y}}_\tau) + D(\y_{\tilde N_\tau}, \hat{\mathbf{y}}_\tau) + S(\y_{\tilde N_\tau}, \hat{\mathbf{y}}_\tau)}
\end{equation}

The motivation behind RWER is to evaluate the quality of the partial transcription at each time point, without penalizing for alignment mismatches.

When transcribing streaming audio, the complete transcription must be produced by the end of the stream, while minimizing per-word latency. Since RWER does not capture latency or alignment, we introduce an additional metric.

\paragraph{Aligned-Relative Word Error Rate (ARWER)} 
Let $\y$ be the ground-truth transcription containing $N$ words, and $\hat{\mathbf{y}}_\tau$ the hypothesis at time $\tau$ with $\tilde{N}_\tau$ words. Define $\hat{\mathcal{Y}} = \{\hat{\mathbf{y}}_0, \hat{\mathbf{y}}_1, \ldots, \hat{\mathbf{y}}_T\}$ as before. Let $\mathbf{y}_\tau$ denote the prefix of the ground-truth that includes only words whose end time is before time $\tau$. Then, 
The ARWER is defined as:
\begin{equation}\nonumber
\mathrm{ARWER}(\mathbf{y},\hat{\mathcal{Y}})\!=\!\frac{\sum_\tau I\left(\mathbf{y}_{\tau},\hat{\mathbf{y}}_\tau\right) + D\left(\mathbf{y}_{_\tau},\hat{\mathbf{y}}_\tau\right) + S\left(\mathbf{y}_{_\tau},\hat{\mathbf{y}}_\tau\right)}{\sum_\tau C\left(\mathbf{y}_{_\tau},\hat{\mathbf{y}}_\tau\right) + D\left(\mathbf{y}_{_\tau},\hat{\mathbf{y}}_\tau\right) + S\left(\mathbf{y}_{_\tau},\hat{\mathbf{y}}_\tau\right)}
\end{equation}
ARWER is better suited for evaluating transcription quality over time, as it considers alignment between the hypothesis and the ground-truth words that are expected to have been transcribed by each point. Models with higher per-word latency will tend to have higher ARWER, due to increased deletions.

\paragraph{Word Error Rate (WER)} We also report the standard WER, calculated between the final hypothesis produced at the end of streaming and the full ground-truth transcription.


\begin{figure*}[ht!]
    \centering
    \subfloat{%
    \includegraphics[width=0.5\textwidth]{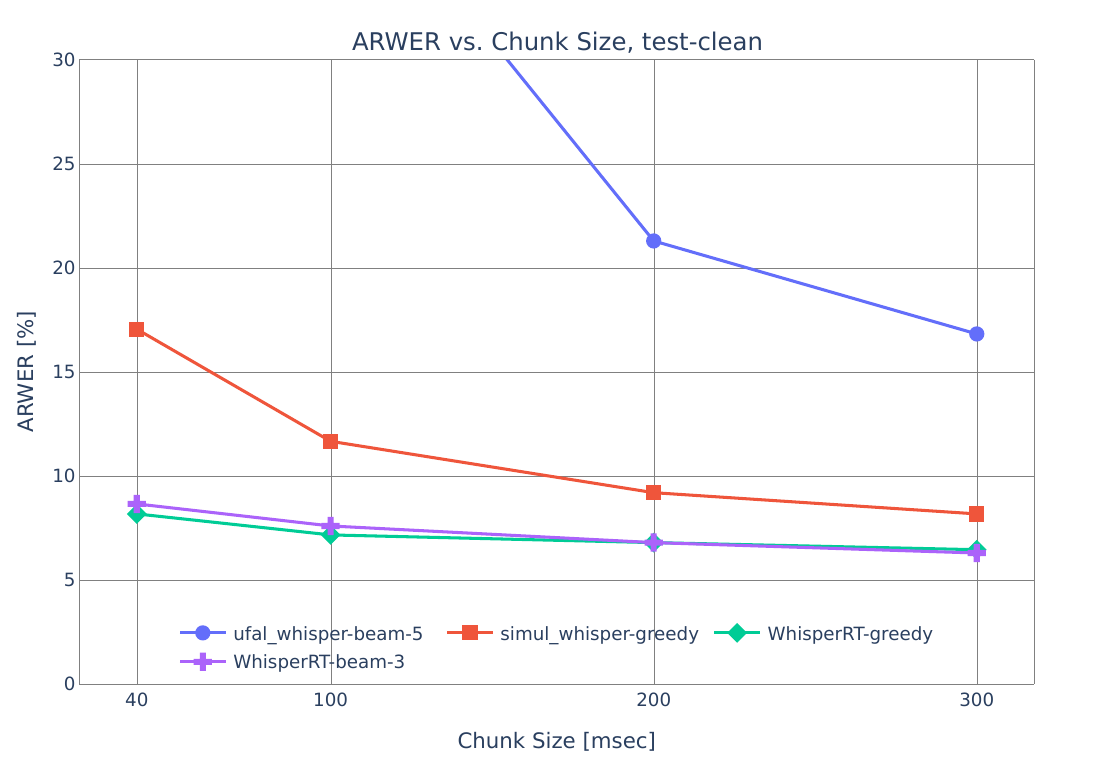}%
    }
    \hfill
    \subfloat{%
    \includegraphics[width=0.5\textwidth]{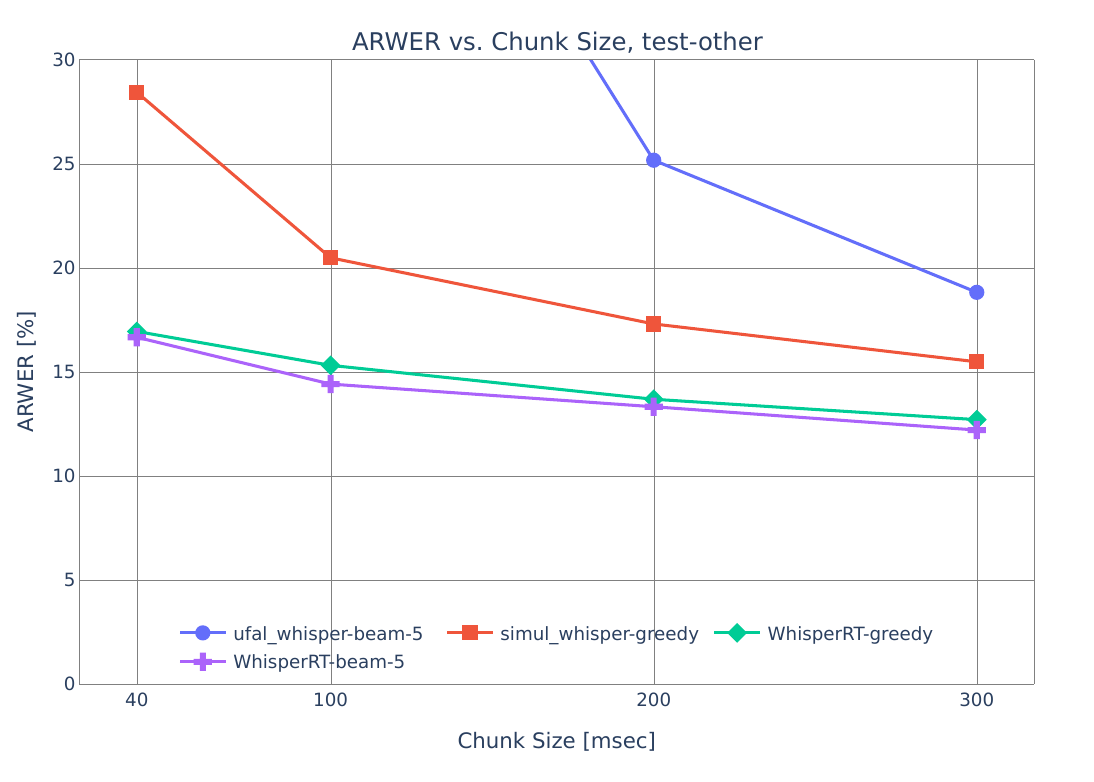}%
    }
    \caption{ARWER vs. Chunk Size per method, on \emph{large-v2} models. Left sub figure presents the results on LibriSpeech test-clean. Right sub figure presents the results on LibriSpeech test-other}
    \label{fig:large-comparison-ARWER}
\end{figure*}

\bigskip

\subsection{English transcription}
For the English transcription task, we fine-tuned three different sizes of the Whisper model: \emph{base}, \emph{small}, and \emph{large-v2}. All models were fine-tuned on the LibriSpeech-train dataset \cite{panayotov2015librispeech}, which consists of 960 hours of read speech. In order to preserve Whisper's punctuations capabilities, we used Librispeech-PC \cite{meister2023librispeechpc} that includes punctuated transcripts. The \emph{base} and \emph{small} models were fine-tuned using LoRA layers with a rank of $r=32$, while the \emph{large-v2} model was fine-tuned with $r=4$. Fine-tuning was conducted using chunk sizes selected from the set $\{40, 100, 200, 300\}$ msec, with an initial chunk size of $600$ milliseconds. We used cross-entropy as the loss function and optimized the models with the AdamW optimizer \cite{loshchilov2019decoupledweightdecayregularization}, using an initial learning rate of $1\text{e}^{-5}$. Full details can be found in the supplementary material.

Evaluation was performed on the LibriSpeech \emph{test-clean} and \emph{test-other} splits, and TED-LIUM3-test split. Each fine-tuned model was evaluated on these datasets using both greedy streaming decoding and streaming beam search with beam size $b=3$ and $n=2$. We report the results for decoding using the beam search setup. We compared our results against \emph{Simul-Whisper} \cite{wang2024simul} and \emph{Ufal-Whisper} \cite{machavcek2023turning}, using the same chunk size settings.

Figure~\ref{fig:large-comparison-ARWER} presents ARWER results across the different methods. The results indicate that our training approach enhances Whisper’s alignment capabilities and outperforms the heuristic approach used in \emph{Simul-Whisper} and \emph{Ufal-Whisper} for both greedy decoding and beam search.


Complete evaluation results are reported in Table~\ref{table:english-transcription-results}. The best scores per metric, per dataset, and per chunk size are marked in bold. Overall, our method consistently outperforms competing approaches across most model sizes on both \emph{test-clean} and \emph{test-other} sets. Our \emph{base} and \emph{small} models seem to have inferior generalization on TED-LIUM3, especially on larger chunk sizes. We suggest that it might be related to the relatively high rank that is used in order to fine tune both models, since our \emph{large-v2} model performs better on the same dataset, using a lower rank.
 
\subsection{Choice of forced aligner} 
Given the variety of available tools for speech dataset alignment, we evaluated the impact of using different forced aligners during training. We compared the Montréal Forced Aligner (MFA) \cite{mcauliffe2017montreal} against Massively Multilingual Speech (MMS) \cite{pratap2023auli}. This comparison was restricted to the \emph{base} and \emph{small} model architectures. Full results are presented in Table \ref{tab:model_comparison_aligners}. Overall, the choice of aligner had a negligible effect on final performance, with both methods yielding comparable results.

\subsection{Training with a random chunk size mask} 
Building upon the discussion in Section \ref{section:training-process}, we trained base, small, and large-v2 models using a random chunk size mask with $\bar f = 30$. Detailed results are provided in Table~\ref{tab:random_masking_model_results}. In general, performance degrades for the 100 ms chunk size compared to models trained with a fixed-size mask. However, for larger chunk sizes, such as 200 ms and 300 ms, the performance gap is minimal or even comparable.

\begin{table*}[ht]
\centering
\begin{tabular}{ll|c|ccc|ccc|cc}
\toprule
\textbf{Model} & \multicolumn{2}{c}{\textbf{System}} & \multicolumn{3}{c}{\textbf{test-clean}} & \multicolumn{3}{c}{\textbf{test-other}} & \multicolumn{2}{c}{\textbf{TED-LIUM3}} \\
\cmidrule(lr){2-3} \cmidrule(lr){4-6} \cmidrule(lr){7-9} \cmidrule(lr){10-11}
& Method & Chunk Size & RWER & ARWER & WER & RWER & ARWER & WER & RWER & WER \\
\midrule

\multirow{13}{*}{\textbf{base}}
  & \multirow{4}{*}{Simul-Whisper \cite{wang2024simul}} & 40  & 27.54  & 30.60  & 25.78  & 42.84  & 48.15  & 42.82 & 23.76 & 24.57     \\ 
  &                        & 100 & 21.60  & 24.82  & 18.27  & 32.26  & 36.15  & 31.49 & \textbf{15.28} & 14.92    \\ 
  &                        & 200 & 13.33  & 16.68  & 12.58  & 27.04  & 30.94  & 26.02 & 12.46 & 12.02    \\ 
  &                        & 300 & 13.52  & 16.86  & 12.79  & 24.27  & 27.85  & 23.46 & 12.46 & 11.40    \\ \cline{2-11}
  & \multirow{4}{*}{Ufal-Whisper \cite{machavcek2023turning}}  & 40  & 48.95  & 60.24  & 49.75  & 57.40  & 71.35  & 55.83     & 37.77 & 36.40 \\ 
  &                        & 100 & 24.01  & 29.26  & 21.43  & 32.37  & 38.63  & 29.43 & 19.62 & 17.81     \\ 
  &                        & 200 & 12.14  & 14.18  & 11.04  & 20.57  & 22.64  & 18.74 & \textbf{12.38} & \textbf{10.63}   \\ 
  &                        & 300 & 8.21  & 9.56  & 7.60 & \textbf{16.78}  & 18.06  & 15.45 & \textbf{9.67} & \textbf{8.14}    \\ \cline{2-11}
  & \multirow{4}{*}{WhisperRT}  & 40  & \textbf{10.57} & \textbf{11.44} & \textbf{9.04} & \textbf{23.19} & \textbf{23.22} & \textbf{20.91} & \textbf{17.56} & \textbf{14.40} \\ 
  &                        & 100 & \textbf{9.28} & \textbf{9.99} & \textbf{7.75} & \textbf{20.36} & \textbf{20.25} & \textbf{17.91} & 15.42 & \textbf{12.49} \\ 
  &                        & 200 & \textbf{8.16} & \textbf{8.85} & \textbf{6.86}  & \textbf{18.61} & \textbf{18.52} & \textbf{16.17} & 13.33 & 10.74  \\ 
  &                        & 300 & \textbf{7.53}  & \textbf{8.12} & \textbf{6.55} & 16.95 & \textbf{16.93} & \textbf{15.18} & 12.72 & 10.46 \\ \cline{2-11} 
  & \multirow{1}{*}{Offline Whisper \cite{radford2022robustspeechrecognitionlargescale}} & - & - & - & 5.00 & - & - & 12.40 & - & 5.00 \\
\midrule

\multirow{13}{*}{\textbf{small}}
  & \multirow{4}{*}{Simul-Whisper \cite{wang2024simul}} & 40  & 17.07  & 21.52  & 14.68  & 29.17  & 35.64  & 32.71 & \textbf{14.53} & 14.18 \\ 
  &                        & 100 & 10.06  & 14.25  & 8.97  & 21.05  & 27.27  & 21.99 & \textbf{10.79} & 10.35 \\ 
  &                        & 200 & 8.03  & 13.22  & 7.58  & 18.02  & 23.41  & 19.26 & \textbf{8.84} & \textbf{7.06} \\ 
  &                        & 300 & 6.88  & 12.21  & 6.43  & 15.69  & 21.66  & 15.76 & \textbf{8.16} & 6.69 \\ \cline{2-11}
  & \multirow{4}{*}{Ufal-Whisper \cite{machavcek2023turning}}  & 40  & 50.20  & 62.01  & 47.95 & 59.90  & 75.70 & 57.07 & 37.08 & 34.11 \\ 
  &                        & 100 & 19.57  & 27.02  & 17.99 & 27.26  & 35.01 & 24.50 & 17.11 & 15.21 \\ 
  &                        & 200 & 8.04  & 11.87 & 7.80 & 14.57 & 18.01 & 13.45 & 10.22 & 8.52 \\ 
  &                        & 300 & 5.68  & 7.40  & 5.42  & \textbf{11.26}  & \textbf{13.06} & \textbf{10.60} & 8.55 & \textbf{6.66} \\ \cline{2-11}
  & \multirow{4}{*}{WhisperRT}  & 40  & \textbf{6.91}  & \textbf{8.59}  & \textbf{5.94}  & \textbf{16.50}  & \textbf{17.40}  & \textbf{14.79} & 14.88 & \textbf{11.91} \\ 
  &                        & 100 & \textbf{6.33}  & \textbf{7.86}  & \textbf{5.54}  & \textbf{15.12}  & \textbf{15.94}  & \textbf{13.68} & 12.24 & \textbf{9.69} \\ 
  &                        & 200 & \textbf{5.61}  & \textbf{6.93}  & \textbf{4.90}  & \textbf{13.23}  & \textbf{14.01}  & \textbf{11.86} & 10.35 & 8.50 \\ 
  &                        & 300 & \textbf{5.39}  & \textbf{6.69}  & \textbf{5.11}  & 12.49  & 13.22  & 11.31 & 9.84 & 7.81 \\ 
  \cline{2-11}
  & \multirow{1}{*}{Offline Whisper \cite{radford2022robustspeechrecognitionlargescale}} & - & - & - & 3.40 & - & - & 7.60 & - & 4.30 \\
\midrule

\multirow{12}{*}{\textbf{large-v2}}
  & \multirow{4}{*}{Simul-Whisper \cite{wang2024simul}} & 40  & 14.59  & 17.05  & 13.63  & 26.07  & 28.44  & 24.62 & 12.75 & 11.64 \\ 
  &                        & 100 & 9.33  &  11.66 & 7.87  & 18.26  & 20.48  & 15.97 & \textbf{9.56} & 9.70 \\ 
  &                        & 200 & 6.62  & 9.19  & 5.61  & 14.65  & 17.30  & 13.24 & \textbf{8.31} & 8.08 \\ 
  &                        & 300 & 5.67  & 8.17  & \textbf{4.57}  & 12.68  & 15.48  & 11.21 & \textbf{8.05} & 7.89 \\ \cline{2-11}
  & \multirow{4}{*}{Ufal-Whisper\cite{machavcek2023turning}}  & 40  & 76.72 & 99.86 & 78.40 & 90.68  & 124.68 & 97.93 & 71.05 & 81.88 \\ 
  &                        & 100 & 32.15 & 40.50 & 27.64 & 40.57  & 49.94  & 35.27 & 30.25 & 30.51  \\ 
  &                        & 200 & 12.78 & 21.29 & 11.26 & 18.00  & 25.17  & 15.70 & 13.65 & 13.31 \\ 
  &                        & 300 & 6.79  & 16.82 & 7.50  & 10.88  & 18.82  & 10.76 & 8.95 & 8.82 \\ \cline{2-11}
  & \multirow{4}{*}{WhisperRT}  & 40  & \textbf{6.33}  & \textbf{8.65}  & \textbf{5.94}  & \textbf{14.74}  & \textbf{16.28}  & \textbf{13.35} & \textbf{11.41} & \textbf{10.38} \\ 
  &                        & 100 & \textbf{5.53}  & \textbf{7.40}  & \textbf{5.07}  & \textbf{12.54}  & \textbf{13.91}  & \textbf{11.17} & 9.80 & \textbf{7.71} \\ 
  &                        & 200 & \textbf{5.02}  & \textbf{6.79}  & \textbf{5.24}  & \textbf{11.73}  & \textbf{12.88}  & \textbf{10.34} & 9.07 & \textbf{7.67} \\ 
  &                        & 300 & \textbf{4.61}  & \textbf{6.29}  & 4.98  & \textbf{10.69}  & \textbf{11.76}  & \textbf{9.57} & 8.70 & \textbf{7.22}  \\
  \cline{2-11}
  & \multirow{1}{*}{Offline Whisper \cite{radford2022robustspeechrecognitionlargescale}} & - & - & - & 2.70 & - & - & 5.20 & - & 3.70 \\
\bottomrule
\end{tabular}
\caption{RWER, ARWER, WER (\%) across models, latency, and metrics. Best results per latency and dataset for each metric are in \textbf{bold}. Whisper results on a non streaming case are added for comparison.}
\label{table:english-transcription-results}
\end{table*}

\bigskip

\begin{table*}[t]
\centering
\begin{adjustbox}{width=\textwidth}
\begin{tabular}{ll|ccc|ccc|ccc|ccc|cc|cc}
\toprule
\multirow{3}{*}{\textbf{Model}} & \multirow{3}{*}{\textbf{Chunk Size}} & \multicolumn{6}{c}{\textbf{test-clean}} & \multicolumn{6}{c}{\textbf{test-other}} & \multicolumn{4}{c}{\textbf{TED-LIUM3}} \\
\cmidrule(lr){3-8} \cmidrule(lr){9-14} \cmidrule(lr){15-18}
& & \multicolumn{3}{c}{\textbf{MFA}} & \multicolumn{3}{c}{\textbf{MMS}} & \multicolumn{3}{c}{\textbf{MFA}} & \multicolumn{3}{c}{\textbf{MMS}} & \multicolumn{2}{c}{\textbf{MFA}} & \multicolumn{2}{c}{\textbf{MMS}} \\
\cmidrule(lr){3-5} \cmidrule(lr){6-8} \cmidrule(lr){9-11} \cmidrule(lr){12-14} \cmidrule(lr){15-16} \cmidrule(lr){17-18}
& & \footnotesize{RWER} & \footnotesize{ARWER} & \footnotesize{WER} & \footnotesize{RWER} & \footnotesize{ARWER} & \footnotesize{WER} & \footnotesize{RWER} & \footnotesize{ARWER} & \footnotesize{WER} & \footnotesize{RWER} & \footnotesize{ARWER} & \footnotesize{WER} & \footnotesize{RWER} & \footnotesize{WER} & \footnotesize{RWER} & \footnotesize{WER} \\
\midrule
\multirow{4}{*}{\textbf{base}} 
 & 40 & 10.57 & 11.44 & 9.04 & 10.68 & 11.53 & 9.07 & 23.19 & 23.22 & 20.91 & 23.22 & 23.15 & 20.61 & 17.56 & 14.40 & 17.80 & 14.51 \\
 & 100 & 9.28 & 9.99 & 7.75 & 9.23 & 10.01 & 7.70 & 20.36 & 20.25 & 17.91 & 20.18 & 20.17 & 18.21 & 15.42 & 12.49 & 14.92 & 12.26 \\
 & 200 & 8.16 & 8.85 & 6.86 & 8.48 & 9.15 & 7.20 & 18.61 & 18.52 & 16.17 & 18.07 & 18.05 & 15.90 & 13.33 & 10.74 & 13.20 & 10.34 \\
 & 300 & 7.53 & 8.12 & 6.55 & 7.84 & 8.51 & 6.85 & 16.95 & 16.93 & 15.18 & 17.03 & 17.05 & 15.14 & 12.72 & 10.46 & 12.54 & 10.16 \\
\midrule
\multirow{4}{*}{\textbf{small}} 
 & 40 & 6.91 & 8.59 & 5.94 & 7.27 & 8.85 & 6.16 & 16.50 & 17.40 & 14.79 & 16.98 & 17.77 & 15.04 & 14.88 & 11.91 & 14.37 & 11.06 \\
 & 100 & 6.33 & 7.86 & 5.54 & 6.32 & 7.77 & 5.47 & 15.12 & 15.94 & 13.68 & 15.41 & 16.14 & 13.60 & 12.24 & 9.37 & 11.89 & 9.18 \\
 & 200 & 5.61 & 6.93 & 4.90 & 5.68 & 7.02 & 5.01 & 13.23 & 14.01 & 11.86 & 13.52 & 14.20 & 11.94 & 10.35 & 8.50 & 10.51 & 8.10 \\
 & 300 & 5.62 & 6.88 & 5.13 & 5.39 & 6.69 & 5.11 & 13.65 & 14.25 & 12.26 & 12.49 & 13.22 & 11.31 & 10.42 & 8.55 & 9.84 & 7.81 \\
\bottomrule
\end{tabular}
\end{adjustbox}
\caption{Comparison of base and small Models: MFA vs. MMS Alignment}
\label{tab:model_comparison_aligners}
\end{table*}

\begin{table*}[ht]
\centering
\begin{tabular}{ll|ccc|ccc|cc}
\toprule
\multirow{3}{*}{\textbf{Model}} & \multirow{3}{*}{\textbf{Gran.}} & \multicolumn{3}{c}{\textbf{test-clean}} & \multicolumn{3}{c}{\textbf{test-other}} & \multicolumn{2}{c}{\textbf{TED-LIUM3}} \\
\cmidrule(lr){3-5} \cmidrule(lr){6-8} \cmidrule(lr){9-10}
& & RWER & A-RWER & WER & RWER & A-RWER & WER & RWER & WER \\
\midrule
\multirow{3}{*}{\textbf{base}} 
 & 100  & 10.58 & 11.26 & 9.00 & 23.56 & 23.42 & 21.51 & 18.71 & 15.38 \\
 & 200 & 8.54 & 9.15 & 7.21 & 18.88 & 18.84 & 16.82 & 14.51 & 11.85 \\
 & 300 & 7.76 & 8.34 & 6.73 & 17.23 & 17.21 & 15.29 & 13.42 & 10.73 \\
\midrule
\multirow{3}{*}{\textbf{small}} 
& 100  & 8.95  & 10.12 & 7.89  & 20.21 & 20.58 & 18.48 & 16.31 & 13.17 \\
 & 200 & 6.72  & 7.98  & 5.91  & 15.66 & 16.21 & 14.01 & 12.94 & 10.40 \\
 & 300 & 6.10  & 7.23  & 5.68  & 14.24 & 14.66 & 12.78 & 11.69 & 9.43  \\
\midrule
\multirow{3}{*}{\textbf{large-v2}} 
 & 100  & 9.92 & 11.59 & 8.92 & 20.48 & 21.17 & 18.27 & 14.68 & 12.94 \\
 & 200 & 6.88 & 8.67 & 6.83 & 15.51 & 16.42 & 13.89 & 10.81 & 9.09 \\
 & 300 & 6.37 & 8.09 & 6.38 & 13.76 & 14.66 & 12.13 & 9.60 & 8.41 \\
\bottomrule
\end{tabular}
\caption{Random masking Model Performance Comparison across Chunks sizes and Datasets}
\label{tab:random_masking_model_results}
\end{table*}

\subsection{Multilingual Transcription}
For the multilingual transcription task, a single \emph{large-v2} model was fine-tuned using a chunk size of $300$ msec and an initial chunk size of $600$ msec. The training data consisted of the Multilingual LibriSpeech corpus (excluding the English portion), supplemented with the full 960-hour LibriSpeech training set for two epochs. Complete details of the fine-tuning are provided in the supplementary material.

Evaluation was conducted on the French, German, Portuguese, and Spanish test subsets of the Multilingual LibriSpeech dataset, using streaming beam search with a beam size of $b=5$ and $n=2$. We compared our model’s performance to that of \emph{Simul-Whisper} and \emph{Ufal-Whisper}, under the same chunking configuration of $300$ msec with a $600$ msec initial chunk.

As shown in Table~\ref{table:mls-transcription-results}, although our LoRA-fine-tuned model outperforms \emph{Simul-Whisper}, it is consistently outperformed by \emph{Ufal-Whisper} in terms of both WER and ARWER across all evaluated languages. This may be due to the limited multilingual exposure during fine-tuning. Unlike the original Whisper model, our fine-tuned version likely does not retain enough linguistic diversity to generalize effectively across multiple languages.

\begin{table*}[t]
\centering
\begin{tabular}{lccccccccccccc}
\toprule
\textbf{Model} 
& \multicolumn{3}{c}{\textbf{French}} 
& \multicolumn{3}{c}{\textbf{German}} 
& \multicolumn{3}{c}{\textbf{Portuguese}} 
& \multicolumn{3}{c}{\textbf{Spanish}} \\
\cmidrule(lr){2-4} \cmidrule(lr){5-7} \cmidrule(lr){8-10} \cmidrule(lr){11-13}
& RWER & ARWER & WER 
& RWER & ARWER & WER 
& RWER & ARWER & WER 
& RWER & ARWER & WER \\
\midrule
Simul-Whisper  \cite{wang2024simul} &   17.4   &  20.65    &  15.24    &   16.20   &   24.79   &   20.40   &  18.44    &  22.15    &  17.16    & 11.58     &  16.03    &  11.64    \\
Ufal-Whisper \cite{machavcek2023turning}   &  16.79    &  \textbf{15.94}    &  \textbf{12.40}    &   \textbf{13.83}   &  \textbf{15.21}    &  \textbf{11.40}    &  \textbf{14.39}    &  \textbf{14.26}    &  \textbf{11.85}    &    13.01  &   \textbf{12.85}   & \textbf{9.98}   \\
WhisperRT    &   \textbf{14.14}   &  17.24    &  14.23    &  14.07    &  17.54    &  14.45    &   18.26   &  20.19    & 16.49    &  \textbf{10.62}    &  13.63    &  10.55    \\
\bottomrule
\end{tabular}
\caption{WER Variants Across Languages, evaluated on \emph{large-v2} models, using a chunk size of $300$ msec, and an initial chunk size of $600$ msec. Best results marked in \textbf{bold}.}
\label{table:mls-transcription-results}
\end{table*}

\subsection{Decoder self-attention KV-cache}
\emph{WhisperRT} enables the use of decoder self-attention KV-caching under certain assumptions. To assess the impact of KV-cache on transcription quality, we conducted an evaluation on the LibriSpeech \emph{test-clean} set, using both greedy streaming decoding and beam search streaming decoding with beam sizes $b \in \{3, 5, 7\}$ and $n=2$.

\begin{figure}
    \centering
    \includegraphics[width=1\linewidth]{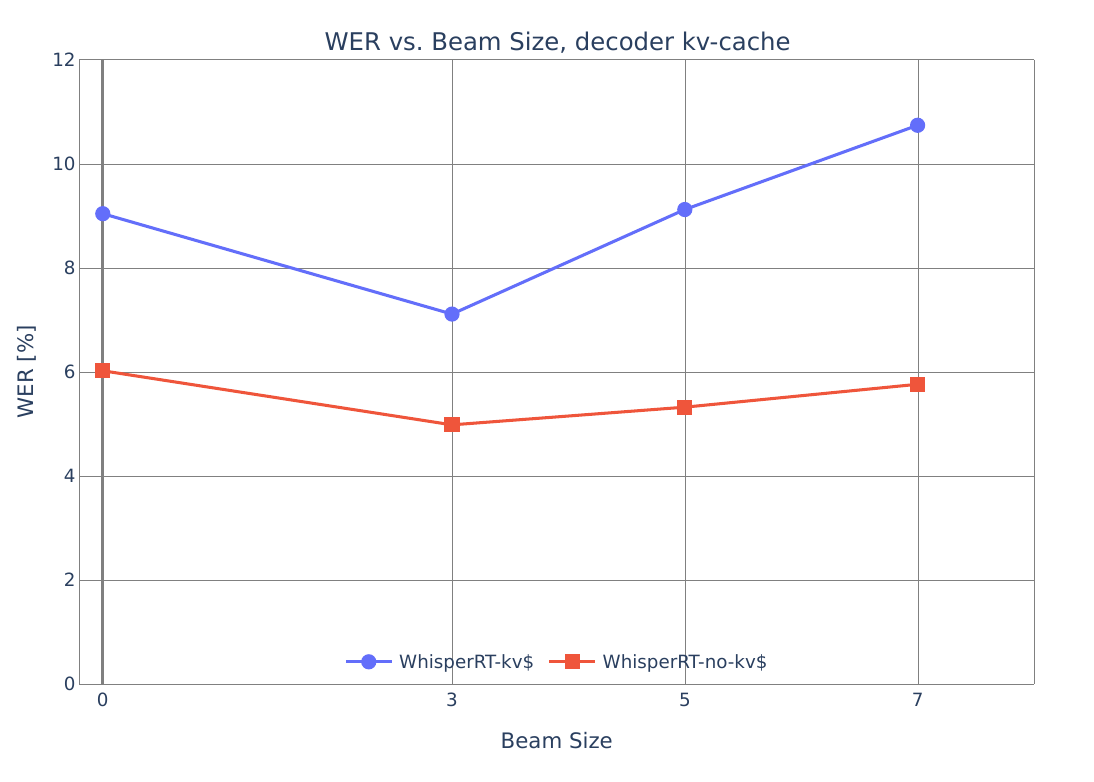}
    \caption{WER vs. beam size on our method when using \emph{large-v2} model. Decoder KV-cache models WER diverges quickly as beam size increases.}
    \label{fig:KV-cache-comparison}
\end{figure}

As shown in Figure~\ref{fig:KV-cache-comparison}, employing KV-cache in the decoder self-attention leads to a significant degradation in transcription performance. Although the encoder output is causal, the cross-attention mechanism appears to benefit from access to newly arriving frames, likely improving token prediction. This aligns with prior findings that not all attention heads function purely as alignment heads.

Future work may address this limitation by introducing regularization mechanisms to reduce the influence of non-alignment heads in cross-attention. Alternatively, a causal design of attention heads could be explored, encouraging stronger alignment between attention and acoustic input.

\subsection{Runtime}

\label{section:experiments-runtime}
We compared the runtime performance of our method against \emph{Simul-Whisper} and \emph{Ufal-Whisper}. During all experiments, we measured the computational latency per frame to compute both the average latency and the Real-Time Factor (RTF) of the model. All evaluations were performed on the LibriSpeech \emph{test-clean} set using a DGX system equipped with 8$\times$ NVIDIA A100 GPUs, each with 80GB of VRAM. To assess runtime efficiency, we used RTF, a common metric for streaming systems. Let $t_c$ be the time required  to process a chunk of size $C$ till an output is emitted. Then, the RTF is defined as:
\begin{equation}
\mathrm{RTF} = \frac{t_c}{C}
\end{equation}

We report the mean RTF across all frames in the stream. Additionally, we report the average per-frame latency, i.e., the average processing time between receiving a new frame and completing its inference.

\begin{figure}
    \centering
    \includegraphics[width=1\linewidth]{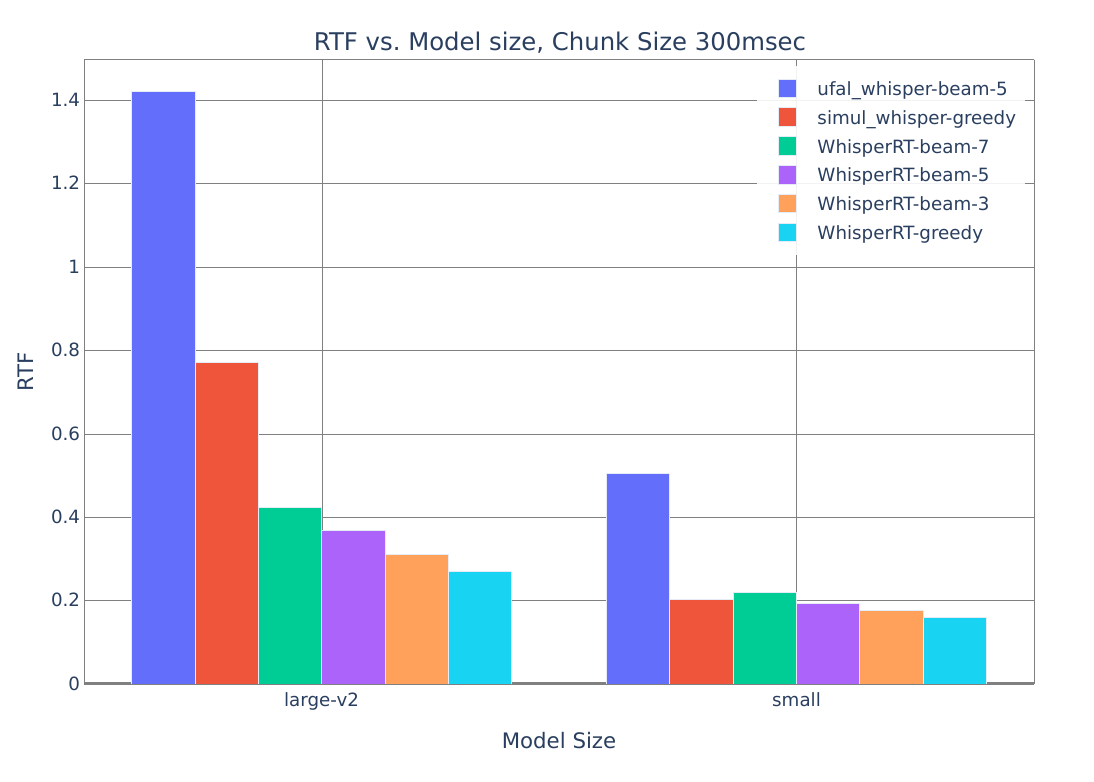}
    \caption{RTF comparison on $300$ msec chunk size streaming case simulation. Note that \emph{Ufal-Whisper} runs faster whisper under the hood, using beam with size 5. \emph{Simul-Whisper} uses greedy decoding.}
    \label{fig:RTF-comp}
\end{figure}

\begin{table}[H]
\centering
\begin{tabular}{lcc}
\toprule
\textbf{Model} & \textbf{Beam Size} & \textbf{Avg. Latency (s)} $\downarrow$ \\
\midrule
Ufal-Whisper \cite{machavcek2023turning}   &   5   &    0.426   \\
Simul-Whisper \cite{wang2024simul}  &   0   &    0.231   \\
WhisperRT   &   0    &   0.081   \\
WhisperRT   &   3    &   0.094    \\
WhisperRT   &   5    &   0.110    \\
WhisperRT   &   7    &   0.127    \\
\bottomrule
\end{tabular}
\caption{Average calculation latency for different models and beam sizes. All models are of size \emph{large-v2} using a chunk size of $300$ msec}
\label{table:average-latency}
\end{table}

As shown in Figure ~\ref{fig:RTF-comp}, our method consistently achieves faster inference times for both the \emph{small} and \emph{large-v2} model sizes. For the \emph{large-v2} model, our method is nearly four times faster than \emph{Ufal-Whisper} (comparing the same beam size, $b=5$), and nearly three times faster than \emph{Simul-Whisper}. For the \emph{small} model, although the performance gap is smaller, our method still outperforms both baselines.

Notably, \emph{Simul-Whisper} relies solely on greedy decoding, while \emph{Ufal-Whisper} employs beam search with $b=5$ and runs on the Faster-Whisper \footnote{\href{https://github.com/SYSTRAN/faster-whisper}{https://github.com/SYSTRAN/faster-whisper}} backend. In contrast, our method uses the original OpenAI Whisper implementation, without any optimizations such as FlashAttention \cite{dao2022flashattentionfastmemoryefficientexact}. None of the compared methods implement decoder-side KV caching. Despite this, our approach remains faster, even when running our proposed streaming beam search with various beam sizes.

Importantly, the benefit of encoder-side KV caching becomes more pronounced with smaller chunk sizes, where the encoder's computational burden increases. Table~\ref{table:average-latency} reports the average per-frame latency for our WhisperRT \emph{large-v2} model with a chunk size of $300$ msec. The empirical results align with the complexity analysis in Theorem~\ref{theorem:encoder-complexity-analysis}. Unlike our method, both \emph{Simul-Whisper} and \emph{Ufal-Whisper} recompute full non-causal encoder self-attention at each step, adding substantial computational overhead. \emph{Ufal-Whisper} incurs particularly high latency and RTF due to its local agreement mechanism, which is computationally expensive. While \emph{Simul-Whisper}'s lightweight cross-attention heuristic yields RTF values closer to ours for \emph{small} models, the non-causal dot product operations result in significantly higher latency for larger models.

\section{Conclusion}
\label{section:conclusions}
We proposed a method for adapting a non-causal transformer ASR model, such as Whisper, into a causal variant that enables significantly faster real-time transcription with minimal performance degradation and reduced computational cost per transcription. We showed that using a random chunk size mask can achieve results which are on par with fixed chunk size models. Leveraging LoRA, our approach supports a dual-setup model that can be dynamically switched based on user requirements, with negligible memory overhead. We also presented a runtime complexity analysis of the proposed method.  However, the method does not yet address the limitations associated with using a decoder KV-cache in streaming scenarios. One possible solution is to mask the cross-attention heads to enforce alignment with the input acoustics, rather than relying on sparsity assumptions in regions temporally distant from the token's origin.  

\bibliographystyle{plain}
\bibliography{references}

\appendices
\section{Proofs}
\label{appendix:proofs}

\begin{property}
Let $k\tau < T$, where $k$ is the frame index and $\tau$ is the chunk size. Denote by $\X_T$, and by $\X_{k\tau}$ the full and the truncated inputs to the encoder, respectively. Let $\Z_T=\mathrm{Encoder}(\X_{T})$ and let $\Z_{k\tau}=\mathrm{Encoder}(\X_{k\tau})$. Then
\setcounter{equation}{27}
\begin{equation}
    [\Z_T]_t \neq [\Z_{k\tau}]_t  ~~~ 1\le t \le k\tau~,
\end{equation}    
where $[\Z_T]_t$ is the $t$-th representation.
\label{theorem:1}
\end{property}
\begin{proof}[Proof of Property 1]
    \label{proof:1}
    Let $\Q_{k\tau}, \K_{k\tau}, \V_{k\tau}$ be the queries, keys, values of $\U_{k\tau}$, and similarly let $\Q_T,\K_T,\V_T$ be the queries, keys and values of $\U_T$.
    By definition:
    \begin{align}
        \mathrm{SA}(\U_{k\tau}) & = \mathrm{Softmax}\left(\frac{{\Q_{k\tau}}{\K_{k\tau}^\top}}{\sqrt{d}}\right) {\V_{k\tau}} \\ 
        \mathrm{SA}(\U_T) & = \mathrm{Softmax}\left(\frac{\Q_{T}\K_{T}^\top}{\sqrt{d}}\right) \V_{T}
    \end{align}
    Let us inspect the $t$-th ($t \le k\tau<T$) row of the results:
    \begin{align}
        \mathrm{SA}(\U_{k\tau})_t & = \frac{\sum_{j=1}^{k\tau}\exp\left(\frac{[{\Q}_{k\tau}]_t^{\top}\cdot [{\K}_{k\tau}]_j}{\sqrt{d}}\right)\cdot [\V_{k\tau}]_j}{\sum_{j=1}^{k\tau}\exp\left(\frac{[{\Q}_{k\tau}]_t^{\top}\cdot [{\K}_{k\tau}]_j}{\sqrt{d}}\right)} \\ 
        \mathrm{SA}(\U_T)_t & = \frac{\sum_{j=1}^{T}\exp\left(\frac{[{\Q}_{T}]_t^{\top}\cdot [{\K}_{T}]_j}{\sqrt{d}}\right)\cdot [\V_T]_j}{\sum_{j=1}^{T}\exp\left(\frac{[{\Q}_{T}]_t^{\top}\cdot [{\K}_{T}]_j}{\sqrt{d}}\right)} \\ 
    \end{align}
    And since $k\tau < T$:
    \begin{equation}
        \mathrm{SA}(\U_{k\tau})_t\neq\mathrm{SA}(\U_T)_t    
    \end{equation}
    Since the encoder function can be described as: 
    \begin{equation}
        \mathrm{Encoder}(\X)=f\left(\W(\mathrm{SA}(\X)+\X)+\mathbf{b}\right)
    \end{equation}
    Since the $\mathrm{SA}$ arguments and outputs are not equal, we can conclude that:
    \begin{equation}
        [\Z_{k\tau}]_t\neq [\Z_T]_t ~~~ \forall 1 \le t \le k\tau
    \end{equation}
\end{proof}




\begin{equation}
\label{eq:causal-mask-definition}
\M_{ij}(k, \tau,\tau_0)=\begin{cases}
    0 & 
    \lceil\frac{i}{\tau}\rceil \geq \lceil\frac{j}{\tau}\rceil
    \lor (i,j)\in [1, \tau_0] \times [1,\tau_0]\\
    -\infty & \text{otherwise}
\end{cases}
\end{equation}

\begin{property}
\label{theorem:2}
Let $k\tau < T$, where $k$ is the chunk index and $\tau$ is the chunk size. Denote by $\X_T$, and by $\X_{k\tau}$ the full and the truncated inputs to the encoder respectively. Let $\tilde{\Z}_T=\mathrm{StreamingEncoder}(\X_T)$ and let $\tilde\Z_{k\tau}=\mathrm{StreamingEncoder}(\X_{k\tau})$. Then
    \begin{equation}
        [\tilde\Z_T]_t=[\tilde\Z_{k\tau}]_t ~~~ 1\le t \le k\tau 
    \end{equation}
    where $[\tilde\Z_T]_t$ is the $t$-th representation.
\end{property}
\begin{proof}[Proof of Property 2]
    \label{proof:2}
    Let $\Q_{k\tau}, \K_{k\tau}, \V_{k\tau}$ be the queries, keys, values of $\U_\tau$, and similarly let $\Q_T,\K_T,\V_T$ be the queries, keys and values of $\U_T$.
    Let us inspect the $t$-th ($t \le k\tau < T$) row of the results:
    \begin{align}
        \widetilde{\mathrm{SA}}(\U_{k\tau})_t & = \frac{\sum_{j=1}^{k\tau}\exp\left(\frac{[\Q_{k\tau}]_t^{\top}\cdot [\K_{k\tau}]_j+\M_{tj}}{\sqrt{d}}\right)\cdot [\V_{k\tau}]_j}{\sum_{j=1}^{k\tau}\exp\left(\frac{[\Q_{k\tau}]_t^{\top}\cdot [\K_{k\tau}]_j+\M_{tj}}{\sqrt{d}}\right)} \\ 
        \widetilde{\mathrm{SA}}(\U_T)_t & = \frac{\sum_{j=1}^{T}\exp\left(\frac{[\Q_T]_t^{\top}\cdot [\K_T]_j+M_{tj}}{\sqrt{d}}\right)\cdot [\V_T]_j}{\sum_{j=1}^{T}\exp\left(\frac{[\Q_T]_t^{\top}\cdot [\K_T]_j+M_{tj}}{\sqrt{d}}\right)} 
    \end{align}
    From the mask construction:
    \begin{align}
        \widetilde{\mathrm{SA}}(\U_{k\tau})_t & = \frac{\sum_{j=1}^{\lceil\frac{t}{m}\rceil}\exp\left(\frac{[\Q_{k\tau}]_t^{\top}\cdot [\K_{k\tau}]_j}{\sqrt{d}}\right)\cdot [\V_{k\tau}]_j}{\sum_{j=1}^{\lceil\frac{t}{m}\rceil}\exp\left(\frac{[\Q_{k\tau}]_t^{\top}\cdot [\K_{k\tau}]_j}{\sqrt{d}}\right)} \\ 
        \widetilde{\mathrm{SA}}(\U_T)_t & = \frac{\sum_{j=1}^{\lceil\frac{t}{m}\rceil}\exp\left(\frac{[\Q_T]_t^{\top}\cdot [\K_T]_j}{\sqrt{d}}\right)\cdot [\V_T]_j}{\sum_{j=1}^{\lceil\frac{t}{m}\rceil}\exp\left(\frac{[\Q_T]_t^{\top}\cdot [\K_T]_j}{\sqrt{d}}\right)} 
    \end{align}
    Since $\X_{k\tau}=[\X_{T}]_1^{k\tau}$:
    \begin{equation}
        [\Q_{k\tau}]_t=[\Q_T]_t,
        [\K_{k\tau}]_{1}^{\lceil\frac{t}{m}\rceil}=[\K_T]_{1}^{\lceil\frac{t}{m}\rceil},
        [\V_{k\tau}]_{1}^{\lceil\frac{t}{m}\rceil}=[\V_T]_{1}^{\lceil\frac{t}{m}\rceil}
    \end{equation}
    Therefore:
    \begin{align}
        \widetilde{\mathrm{SA}}(\U_{k\tau})_t=\widetilde{\mathrm{SA}}(\U_T)_t \ \  \forall t \leq k\tau
    \end{align}
    Since the encoder function can be described as: 
    \begin{equation}
        \mathrm{StreamingEncoder}(\U)=f\left(\W(\widetilde{\mathrm{SA}}(\U)+\U) + \mathbf{b}\right)
    \end{equation}
    And the $\widetilde{\mathrm{SA}}$ arguments and outputs are equal, we can conclude that:
    \begin{equation}    
        [\tilde\Z_{k\tau}]_t=[\tilde\Z_T]_t \ \ \forall t \leq k\tau
    \end{equation}
\end{proof}

\begin{claim}[Streaming Greedy Decoding Optimality]
Let 
\begin{equation}
P^*(y_i\mid \y_{<i}, \X_{k\tau})
\end{equation}
be the optimal probability distribution of the $i$-th token given the preceding tokens, and the acoustic data till $k\tau$. Denote by $\rho_k$ the \emph{probability path} of the predicted token sequence until the chunk $k$. Our proposed greedy decoding algorithm, as depicted on Algo.~\ref{alg:greedy-decoding}, is optimal in the sense that it yields a probability path $\rho_k^{\text{CW}}$ which is larger or equal on the probability path of the greedy algorithm, $\rho_k^{\text{G}}$.
\end{claim}
\begin{proof}[Proof of Claim 1]
We prove the claim using mathematical induction. For the first chunk $k=1$ both algorithms choose the same tokens hence have same probability path:
\begin{equation}
\rho_1^{\text{CW}}\geq\rho_1^G
\end{equation}
Assuming the statement is correct for any $k$, we will prove that it holds for $k+1$. For the case of $k$, the greedy decoder continues its prediction from where it previously stopped. The algorithm needs to determine how to handle the token at index $i - m$ considering the options:
1) If $y_{i-m} = j$ is unstable, it means: 
\begin{multline}
    P(y_{i-m} = j \mid \y_{<i-m}, \X_{(k+1)\tau}) \\ < \max_{v\in\mathcal V} P(y_{i-m}\mid  \y_{<i-m}, \X_{(k+1)\tau})
\end{multline}
and
\begin{multline}
    P(y_{i-m} = j \mid \y_{<i-m}, \X_{(k+1)\tau}) \\
    < P(y_{i-m} = j \mid \y_{<i-m}, \X_{k\tau})
\end{multline}
2) If $y_{i-m} = j$ is stable, either:
\begin{equation}
    j = \arg\max_{u \in \mathcal{V}} P(y_{i-m} = u \mid \y_{<{i-m}}, \X_{k\tau}) ~.
\end{equation} 
or 
\begin{equation}
    P(y_{i-m} = j \mid \y_{<{i-m}}, \X_{k\tau}) \ge P(y_{i-m} = j \mid \y_{<{i-m}}, \X_{(k-1)\tau}) 
\end{equation}
holds. Either way, $y_{i-m}$ token is a token with higher probability than the last frame. Thus, $\rho_{k+1}^{\text{CW}}\geq\rho_{k+1}^G$
    
\end{proof}

\begin{theorem}
\label{theorem:4}
Let $T$ be the input sequence length to the encoder, $d$ the embedding dimension, and $\tau$ the chunk size, with $0 < \tau \ll T$. The computation of blocked causal attention over the full sequence during streaming requires $\mathcal{O}(T^2d + Td^2)$ operations and $\mathcal{O}(Td)$ additional memory.
\end{theorem}
\begin{proof}[Proof of theorem 1]
    \label{proof:4}
    We break down the attention calculation process. We observe a process of an arbitrary chunk during the streaming process, let it be chunk $i, ~ 1\le i\le \frac{T}{\tau}$. First, the encoder calculates the projections $\Q, \K, \V \in \R^{\tau\times d}$ of the received chunk. Such calculation requires $\mathcal{O}(\tau d^2)$ operations. In addition, an extra $\mathcal{O}(\tau d)$ memory is required to cache $\K,\V$ of the received chunk. For the dot product calculation, each chunk attends only to previous chunks. Thus, the $i$-th chunk, requires $\mathcal{O}(i\tau^2d)$ operations to calculate the dot product results. In order to calculate the number of operations required to calculate the causal encoder output sequence of length $T$, we sum over the operations per chunk. For the projections calculation: 
    \begin{equation}
        \sum_{i=1}^{\frac{T}{\tau}} \tau d^2 = \frac{T}{\tau}\cdot \tau d^2 = Td^2
    \end{equation}
    Overall, $\mathcal{O}(Td^2)$ operations required.
    When summing over the required operations for the dot products:
    \begin{align}
        \sum_{i=1}^{\frac{T}{\tau}}i\tau^2d& =\frac{\tau^2d}{2}(\frac{T^2}{\tau^2}+\frac{T}{\tau}) \\
        & = \frac{T^2}{2}\cdot d + \frac{T}{2}\cdot \tau d
    \end{align}
    Overall, $\mathcal{O}(T^2d)$ operations required. For the memory complexity, we recall that $\mathcal{O}(\tau d)$ is the memory required per chunk, given $\frac{T}{\tau}$ chunks, we get $\mathcal{O}(Td)$ extra memory required.
\end{proof}

\section{Training Details}
\setcounter{table}{6}

\begin{table}[ht]
\centering
\caption{Fine-tuning parameters for each model variant and chunk size}
\label{tab:training_params}
\renewcommand{\arraystretch}{1.2}
\begin{tabular}{llccc}
\toprule
\textbf{Model} & \textbf{Chunk Size (ms)} & \textbf{Rank} & \textbf{Extra Chunks} & \textbf{Points Fraction} $\hat f$ \\
\midrule
\multirow{4}{*}{Base} 
  & 40     & 32 & 14 & 0.02   \\
  & 100    & 32 & 5 & 0.05   \\
  & 200    & 32 & 2 & 0.10   \\
  & 300    & 32 & 1 & 0.25   \\
\midrule
\multirow{5}{*}{Small} 
  & 40     & 32 & 14 & 0.02   \\
  & 100    & 32 & 5 & 0.05   \\
  & 200    & 32 & 2 & 0.10   \\
  & 300    & 32 & 1 & 0.25   \\
  & 1000   & 32 & 0 & 0.40   \\
\midrule
\multirow{5}{*}{Large-v2} 
  & 40     & 4 & 14 & 0.02   \\
  & 100    & 4 & 5 & 0.05   \\
  & 200    & 4 & 2 & 0.07   \\
  & 300    & 4 & 1 & 0.1   \\
  & 1000   & 4 & 0 & 0.3   \\
\bottomrule
\end{tabular}
\end{table}

\label{appendix:training-details}
\subsection{English Transcription}
We fine-tuned our model using 960 hours of Librispeech-train dataset. All of the models variants were trained with an initial learning rate of 1e-5, using an AdamW \cite{loshchilov2019decoupledweightdecayregularization} optimizer, weight decay of 0.01 and a \emph{ReduceLROnPlateau} learning rate scheduler with factor of 0.5 and patience of 2 epochs. The \emph{base} and \emph{small} variants were trained using batch size of 32 and 16 respectively. Both models were trained for 10 epochs. The \emph{large-v2} models were trained using a batch size of 4 and for 3 epochs. Rest of the hyper parameters are detailed on Table~\ref{tab:training_params}. We used PyTorch Lightning framework to accelerate the training process, using DDP \cite{li2020pytorchdistributedexperiencesaccelerating} strategy.

\subsection{Multilingual Transcription}
We fine-tuned a single \emph{large-v2} model on a Multilingual LibriSpeech (MLS) corpus, excluding the english partition, combined with the 960 hours of LibriSpeech-train. Model was trained using an initial learning rate of 1e-5, using an AdamW optimizer, weight decay of 0.01 for 2 epochs. We used a LoRA rank of 4, with points fraction $\hat f = 0.05$.
\end{document}